\documentclass[iicol,pdflatex,sn-mathphys-num]{sn-jnl}

\usepackage{graphicx}%
\usepackage{multirow}%
\usepackage{amsmath,amssymb,amsfonts}%
\usepackage{amsthm}%
\usepackage{mathrsfs}%
\usepackage[title]{appendix}%
\usepackage[table]{xcolor}%
\usepackage{textcomp}%
\usepackage{manyfoot}%
\usepackage{booktabs}%
\usepackage{makecell}%
\usepackage{pifont}%
\usepackage{caption}%
\usepackage{xspace}%
\usepackage[normalem]{ulem}%
\usepackage{algorithm}%
\usepackage{algorithmicx}%
\usepackage{algpseudocode}%
\usepackage{listings}%
\usepackage{placeins}%

\theoremstyle{thmstyleone}%

\theoremstyle{thmstyletwo}%

\theoremstyle{thmstylethree}%
\newcommand{\cmark}{\textcolor{green!60!black}{\ding{51}}}
\newcommand{\xmark}{\textcolor{red!75!black}{\ding{55}}}

\newcommand{\methodtable}{C\textsuperscript{2}E-S2SER}
\DeclareRobustCommand{\method}{\texorpdfstring{C\textsuperscript{2}E-S2SER}{C2E-S2SER}\xspace}
\let\latexunderline\underline
\renewcommand{\underline}[1]{\uline{#1}}

\begin{document}

\title[Article Title]{When Recovery Matters: The Blind Spot of Surrogate Privacy in MLLM Editing}

\author[1]{\fnm{Siyuan} \sur{Xu}}\email{siyuanxu333@gmail.com}\equalcont{These authors contributed equally to this work.}
\author[1]{\fnm{Yibing} \sur{Liu}}\email{lyibing112@gmail.com}\equalcont{These authors contributed equally to this work.}
\author[1]{\fnm{Peilin} \sur{Chen}}\email{plchen3@cityu.edu.hk}
\author[2]{\fnm{Yung-Hui} \sur{Li}}\email{yunghui.li@foxconn.com}
\author[1]{\fnm{Shiqi} \sur{Wang}}\email{shiqwang@cityu.edu.hk}
\author[3]{\fnm{Sam} \sur{Kwong}}\email{samkwong@ln.edu.hk}

\affil[1]{\orgname{City University of Hong Kong}, \orgaddress{\city{Hong Kong}}}
\affil[2]{\orgname{Hon Hai Research Institute}, \orgaddress{\city{Taipei}}}
\affil[3]{\orgname{Lingnan University}, \orgaddress{\city{Hong Kong}}}

\abstract{
Multimodal Large Language Models (MLLMs) enable flexible instruction-driven image editing, but privacy risks arise when user images expose diverse and user-specific private content.
Canonical privacy protection strategies typically substitute sensitive regions with surrogate content before cloud editing. Yet, the resulting output is often an edited surrogate rather than the desired edited source image, neglecting the local recovery in both design and evaluation scope.
To this end, we introduce SPPE (Surrogate-based Privacy-Preserving Editing), the first recovery-oriented benchmark covering 36 fine-grained privacy categories and 65 editing instructions.
It defines two complementary tasks: 1) editability assessment, which estimates before cloud interaction whether a surrogate can induce an edit consistent with the original image; and 2) surrogate-to-source edit recovery, which evaluates whether the edited surrogate can be transferred back to the private source with the edit effect preserved.
We address each task with a dedicated method: ERMA predicts surrogate editability through instruction-aware multimodal relation modeling, while \method performs cycle-consistent recovery by using the surrogate editing pair as visual edit evidence and the source image as a source-preserving anchor.
Experiments on SPPE and InstructPix2Pix show consistent improvements on both tasks.
For editability assessment, ERMA improves over the best-performing baselines by 13.9\% in SRCC and 12.3\% in PLCC.
For surrogate-to-source edit recovery, \method outperforms SOER across all 8 source integrity and edit consistency metrics on SPPE.
}


\maketitle

\section{Introduction}

\begin{figure*}[t]
    \centering
    \includegraphics[width=1\linewidth]{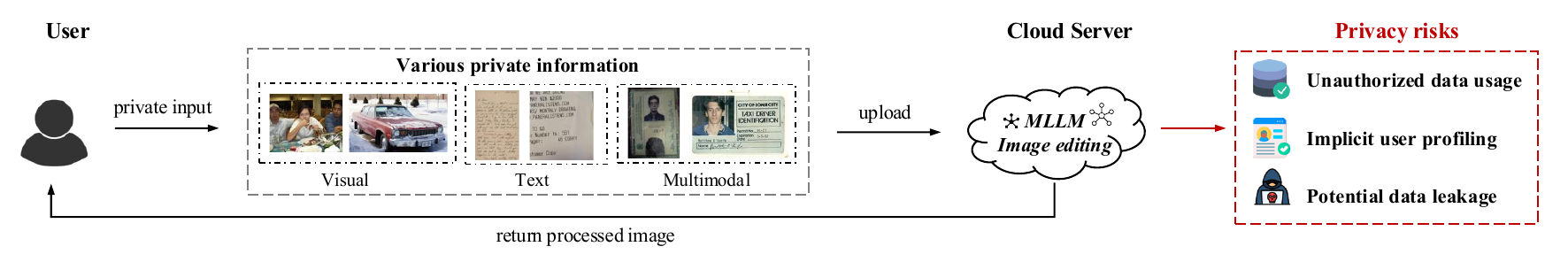}
    \caption{Privacy leakage risks in MLLM image editing. Uploading a user image to a cloud-based MLLM exposes sensitive visual, textual, and multimodal content to unauthorized data usage, implicit profiling, and potential leakage.}
    \label{fig:problem}
\vspace{-12pt}
\end{figure*}
\begin{figure*}[t]
    \centering
    \includegraphics[width=1\linewidth]{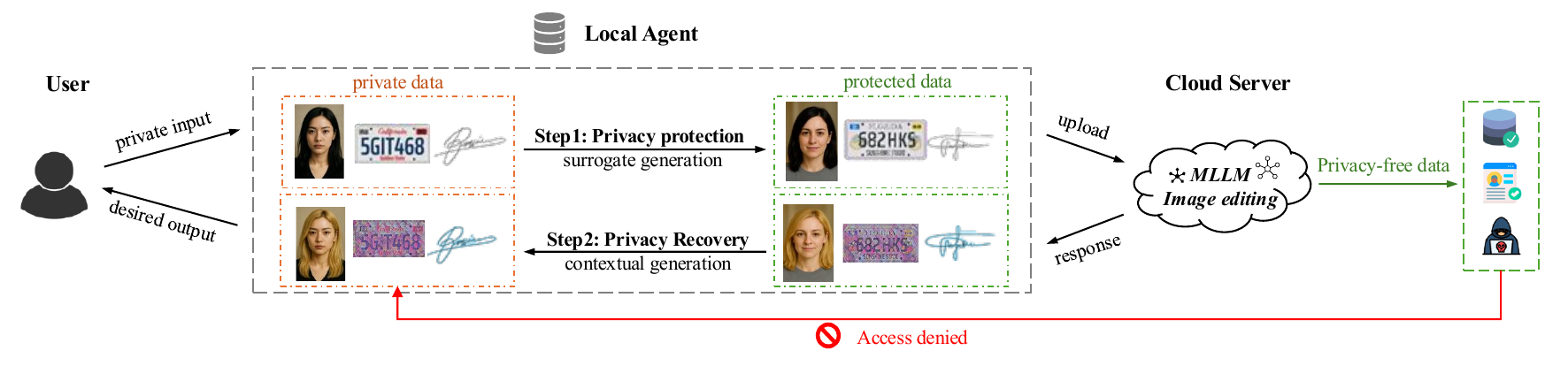}
    \caption{Surrogate-driven MLLM image editing pipeline. A local agent first generates a surrogate by replacing selected sensitive regions with synthesized content. The cloud-hosted MLLM edits only the surrogate, and a local recovery module then transfers the observed editing transformation back to the original image.}
    \label{fig:pipeline}
\vspace{-14pt}
\end{figure*}
Multimodal Large Language Models (MLLMs) have recently demonstrated remarkable capabilities in instruction-guided image editing~\cite{emu,smartedit,x2edit,mgedit}. In these scenarios, user-provided images are often uploaded to cloud services, exposing sensitive content such as faces, documents, or private environments, as illustrated in Fig.~\ref{fig:problem}.

To mitigate such exposure, existing privacy-preserving methods mainly focus on input-side protection. As shown in Fig.~\ref{fig:pipeline}, they typically construct privacy-preserving surrogates by masking, redacting, perturbing, or substituting sensitive regions before cloud processing for privacy protection~\cite{revision,surrogate1,surrogate2,surrogate3,CHI}. However, privacy-preserving image editing usually introduces an output-side challenge, which unexpectedly contradicts user expectations -- \textit{delivering intact editing quality yet ensuring genuine privacy protection}.
Specifically, since the cloud model edits the protected surrogate rather than the original image, local recovery is often made necessary by transferring the observed edit from the edited surrogate back to the source image locally. Nevertheless, a sound mechanism that bridges the gap between surrogate-based editing and lossless local recovery still remains under-explored.

Given the limited evaluation scope of current privacy-preserving MLLM studies~\cite{revision,surrogate1,surrogate2,surrogate3,CHI}, in this paper, we firstly introduce \textbf{SPPE} (\textbf{S}urrogate-based \textbf{P}rivacy-\textbf{P}reserving \textbf{E}diting) benchmark for surrogate-based privacy-preserving image editing. SPPE covers 36 fine-grained privacy categories and 65 diverse editing instructions, spanning diverse style, appearance, object, background, and semantic transformations. SPPE defines two tasks:
(1) \textbf{Editability Assessment}: determine whether a surrogate preserves the edit-relevant semantics needed for downstream editing.
(2) \textbf{Surrogate-to-Source Recovery}: reconstruct the edited original image from the edited surrogate and the private source image.
Together, they capture the two core challenges of recovery-oriented surrogate editing: pre-edit feasibility assessment of the surrogate and post-edit surrogate-to-source edit recovery.

 We further propose \textbf{ERMA} (\textbf{E}ditability-aware \textbf{R}elational \textbf{M}ulti-modal \textbf{A}ssessment) for editability assessment, and \textbf{\method} (\textbf{C}ycle-\textbf{C}onsistent \textbf{E}dit-Conditioned \textbf{S}urrogate-to-\textbf{S}ource \textbf{E}dit \textbf{R}ecovery) for surrogate-to-source edit recovery.
 On one hand, ERMA leverages multi-modal relational modeling to predict surrogate editability by capturing whether relevant semantics are preserved in the surrogate relative to the editing instruction. On the other hand, \method uses a diffusion transformer with visual reference conditioning to transfer the transformation observed in the surrogate pair back onto the source image. \method incorporates edit-conditioned tag generation to infer the actual transformation from the surrogate editing pair, addressing the ambiguity that the same text instruction may produce different visual effects on different surrogates and thereby strengthening edit consistency during recovery. It also uses cycle-consistent regularization through a reverse recovery path that imposes a source-reconstruction constraint on the recovered result, helping reduce unnecessary source drift and improve source integrity. Experiments on SPPE show that ERMA outperforms all quality assessment baselines on editability prediction, and \method achieves consistent improvements over the best recovery baseline across all overall source integrity and edit consistency metrics.
Our main contributions are summarized as follows:
\begin{itemize}
  \item We introduce \textbf{SPPE}, a recovery-oriented benchmark for surrogate-based MLLM privacy-preserving image editing, covering 36 fine-grained privacy categories and 65 editing instructions with two complementary tasks: editability assessment and surrogate-to-source recovery.
  \item We propose \textbf{ERMA} for instruction-aware surrogate editability assessment via multi-modal relational modeling, outperforming the best quality assessment baseline by 13.9\% in SRCC and 12.3\% in PLCC on SPPE.
  \item We develop \textbf{\method}, a diffusion-transformer recovery model that transfers the edit observed in the surrogate pair back to the private source with edit-conditioned guidance and cycle-consistent regularization. It outperforms all recovery baselines across all 8 overall source integrity and edit consistency metrics on SPPE and further shows strong generalization on InstructPix2Pix.
\end{itemize}

\section{Related work}

\subsection{Privacy in Multimodal Large Language Models}
Multimodal Large Language Models (MLLMs) have demonstrated remarkable abilities in integrating and reasoning over multiple modalities. Approaches such as LLaVA and MiniGPT-4~\cite{llava, minigpt4} build on LLaMA~\cite{llama} with instruction tuning, achieving strong results in tasks including visual question answering, grounded reasoning, and interactive human-AI scenarios. To enable image generation, GILL~\cite{gill} combines LLMs with diffusion models, while SEED and SEED-2~\cite{seed, seed2} introduce visual tokenizers to align image and text embeddings for coherent outputs. More recently, SmartEdit~\cite{smartedit} incorporates a Bidirectional Interaction Module to improve instruction comprehension and editing accuracy. These developments illustrate the ongoing trend toward generalist MLLMs capable of both understanding and manipulating visual content. Existing privacy-preserving strategies for MLLMs include Differential Privacy ($\text{DP}$)\cite{dp1,dualpriv} and inference-time protection mechanisms such as ReVision~\cite{revision}. While these methods are effective for general-purpose protection, they are limited in scenarios that require direct visual manipulation, such as image editing, since they do not operate on concrete visual inputs. Surrogate-based approaches~\cite{surrogate1,surrogate2,surrogate3,CHI} mitigate this by replacing sensitive regions with synthetic substitutes, yet they often prioritize privacy at the expense of content consistency, leading to outputs that may deviate from the intended edits. As a result, the problem of faithfully recovering edited results in diverse and flexible MLLM editing scenarios remains largely unaddressed.

\subsection{Visual In-context Learning}
Visual references can convey layouts and attributes that are difficult to specify with text alone, motivating reference-based image editing methods~\cite{zhang2021multi,huang2021unsupervised,cross_image,zhou2025attention,yang2023paint,chen2023specref,he2025freeedit,chen2024zero,biswas2025pixels,chen2024anydoor}. These methods leverage auxiliary images to guide editing or attribute transfer, but many of them are optimized for appearance-level correspondence and remain limited for complex semantic or structural changes, such as identity-consistent edits, localized object removal, or large background transformations. In-context learning has also been extended from language~\cite{brown2020language} to vision, where models infer task patterns from visual exemplars. Existing studies cover visual understanding~\cite{wang2023images,bar2022visual,zhang2023makes,bai2024sequential} and generative settings; for example, IC-LoRA~\cite{iclora} shows that diffusion transformers can synthesize outputs conditioned on visual exemplars. Our setting shares the intuition that an example pair can reveal an editing transformation, but differs in its recovery objective. Surrogate-to-source recovery requires one-to-one alignment between each surrogate editing pair and the private source, emphasizing faithful content recovery and precise transfer of MLLM-specific edit effects. This specificity makes in-context conditioning indispensable in our privacy-preserving scenario: unlike standard editing tasks where a text prompt alone can specify the desired output, the actual visual transformation executed by the cloud MLLM is only observable through the surrogate editing pair. The in-context pair provides critical cues that cannot be inferred from the instruction alone.
\begin{figure*}[t]
\centering
\includegraphics[width=1\linewidth]{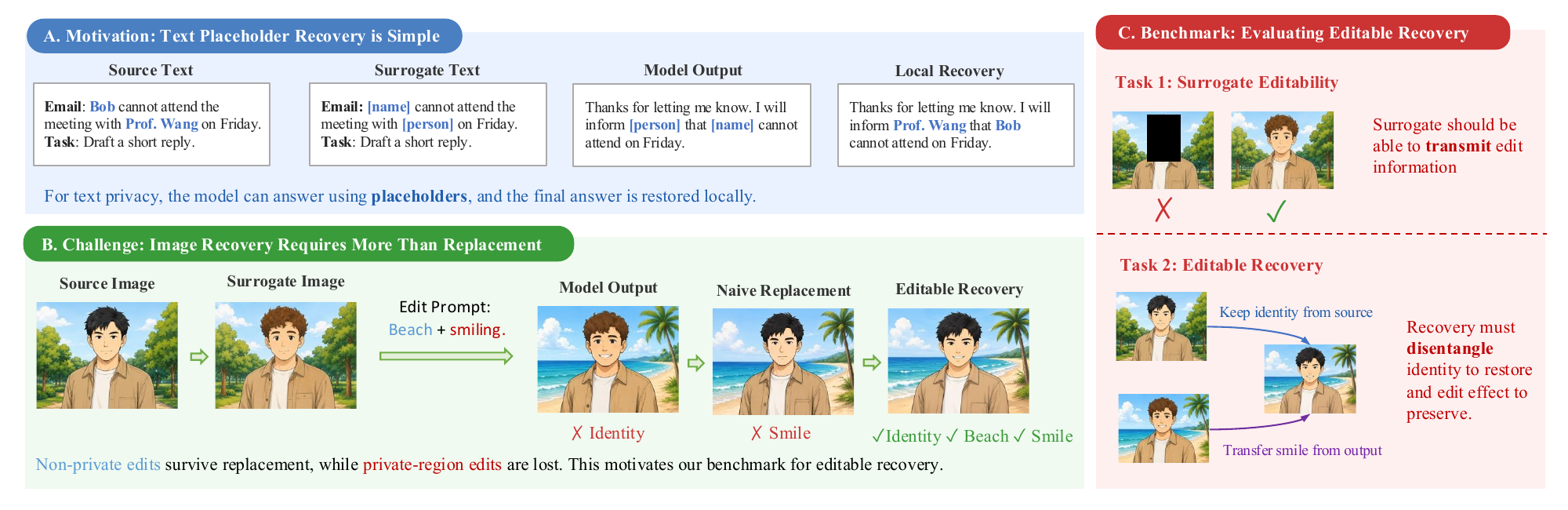}
\caption{Motivation and task design of SPPE. Text placeholders can often be restored by direct replacement, whereas visual surrogate recovery cannot simply replace content back: it must transfer the edit applied to the surrogate while preserving source-specific content. SPPE therefore evaluates two tasks: surrogate editability assessment and surrogate-to-source edit recovery.}

\vspace{-20pt}
\label{fig:bench}

\end{figure*}

\begin{table*}[t]
\raggedright
\resizebox{\textwidth}{!}{%
\begin{tabular}{l|ccc|cc|cc|c}
\toprule
\multirow{2}{*}{\textbf{Dataset}} & \multicolumn{3}{c|}{\textbf{Annotation}} & \multicolumn{2}{c|}{\textbf{Application}} & \multicolumn{2}{c|}{\textbf{Evaluation}} & \multirow{2}{*}{\textbf{Data Scale}} \\

 & {Category} & {Location} &{Anno. Type} & {Task} &{MLLM} & Utility & Recovery & \\

\hline
VISPR \cite{vispr}     &  \cmark   & \xmark &    label &   Understanding  & \xmark & \xmark & \xmark &  12,000  \\
VizWiz-Priv\cite{vizwiz_priv} & \cmark & \cmark &    label,segmentation &  Detection & \xmark & \xmark & \xmark &  5,537   \\
Redactions\cite{visual_redaction} & \cmark & \cmark    &  label,segmentation   &  Detection & \xmark & \xmark & \xmark & 8,473    \\
DIPA\cite{dipa}  & \cmark & \xmark &   label,bbox  &  Detection & \xmark & \xmark & \xmark &   1,495  \\
DIPA2\cite{dipa2}     & \cmark & \xmark &  label,bbox   &  Detection & \xmark & \xmark & \xmark &   1,304  \\
BIV-Priv-Seg\cite{biv_priv} & \cmark &   \cmark  &  label,segmentation   &  Detection & \xmark & \xmark & \xmark &    728* \\
Multi-P2A\cite{multip2a}  & \cmark   & \xmark &  label   &  Assessment & \cmark & \cmark & \xmark  &   31,962 \\
ReVision\cite{revision} & \cmark & \xmark & label& Protection& \cmark &\cmark &\xmark &1,700 \\
HR-VISPR\cite{hrvispr} &\cmark &\cmark & label, segmentation & Assessment& \xmark & \xmark & \xmark & 10,110\\

\hline

\textbf{Ours} & \cmark & \cmark & label, segmentation, bbox & Image Editing &\makecell[c]{ \cmark(65 prompts)} & \cmark & \cmark  &    55,696 \\
\bottomrule

\end{tabular}
}
\caption{Benchmark comparison with existing privacy datasets. \textbf{Category} indicates whether privacy categories are annotated; \textbf{Location} indicates whether sensitive region localization annotations are provided; \textbf{Anno. Type} lists the annotation format; \textbf{Task} summarizes the main application; \textbf{MLLM} indicates whether the dataset is related to MLLM downstream tasks; \textbf{Utility} and \textbf{Recovery} indicate whether the dataset supports evaluating downstream task utility and post-processing recovery, respectively.
}
\label{tab:dataset}
\vspace{-15pt}
\end{table*}

\subsection{Image Quality Assessment}
Image quality assessment (IQA) aims to predict perceptual image quality and has been studied under different assumptions and objectives. Full-reference IQA (FR-IQA) compares a distorted image with a reference image, ranging from hand-crafted fidelity measures such as SSIM~\cite{ssim} and FSIM~\cite{fsim} to learned perceptual or semantic similarity metrics such as DISTS~\cite{dists}, ST-LPIPS~\cite{STLPIPS}, TOPIQ~\cite{TOPIQ}, and AFINE-FR~\cite{AFINE-FR}. These methods are useful when a reference image is available, but their quality notion is usually tied to distortion visibility, structural preservation, or perceptual similarity. In contrast, no-reference IQA (NR-IQA) predicts quality from a single image without a pristine reference. Classical NR-IQA methods such as BRISQUE~\cite{brisque} and NIQE~\cite{niqe} rely on natural scene statistics, while learning-based methods such as NIMA~\cite{nima}, MUSIQ~\cite{musiq}, AHIQ~\cite{ahiq}, and MANIQA~\cite{maniqa} model technical quality, aesthetic preference, or complex synthetic distortions with deep networks. Related reconstruction-based anomaly detection methods also model local visual deviations by reconstructing normal feature patterns and identifying inconsistent regions, e.g., neighborhood-attention-based feature reconstruction~\cite{xian2026neighborhood}. However, these methods are designed for detecting abnormal regions rather than assessing whether a privacy-preserving surrogate retains instruction-relevant editable semantics. These studies collectively focus on predicting how visually faithful, aesthetically pleasing, or perceptually similar an image is, with the quality criterion defined in a task-agnostic manner. In privacy-preserving image editing, however, what matters is whether a protected surrogate preserves the specific visual semantics required by a downstream MLLM editing instruction. This criterion is not captured by existing IQA methods, which do not condition on the editing intent and are not designed to evaluate downstream task utility under privacy constraints.

\section{SPPE: Surrogate-based Privacy-Preserving Editing Dataset}
\subsection{Overview}
We introduce \textbf{SPPE} (\textbf{S}urrogate-based \textbf{P}rivacy-\textbf{P}reserving \textbf{E}diting), a benchmark dataset for privacy-preserving MLLM image editing. SPPE follows a surrogate-driven paradigm in which private content is replaced locally, the surrogate is edited by an MLLM, and the edited result is recovered on the original image. Fig.~\ref{fig:bench} illustrates why recovery under image editing requires dedicated benchmark:  unlike text placeholders that can often be directly restored after processing, visual surrogate recovery cannot rely on direct replacement, since it must transfer the edit effect observed on the surrogate back to the source image while preserving source-specific details. Table~\ref{tab:dataset} summarizes its comparison with existing privacy datasets: SPPE covers 36 fine-grained privacy categories, 65 editing instructions, and 55,696 editing instances. The broad privacy taxonomy provides diverse sensitive scenarios, while the varied editing instructions cover diverse MLLM editing behaviors.

\subsection{Data Format}
\begin{figure*}[t]
    \centering
    \includegraphics[width=1\linewidth]{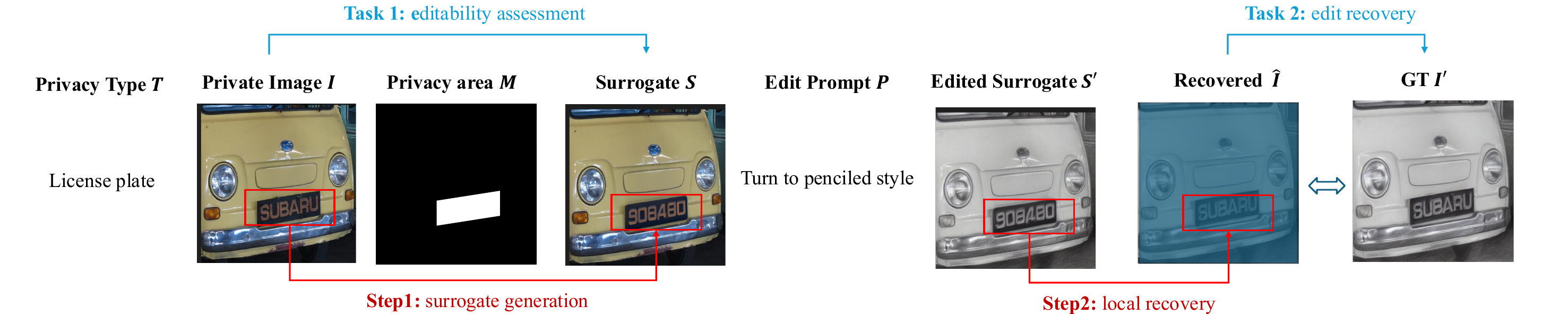}
    \caption{Illustration of an SPPE data instance for privacy-preserving MLLM image editing. Each instance contains a private source image, privacy annotation, surrogate image, editing instruction, edited surrogate, and ground-truth edited source; the recovered result is produced by a local recovery model for evaluation.}
    \label{fig:sample}
    \vspace{-18pt}
\end{figure*}
As illustrated in Fig.~\ref{fig:sample}, each SPPE instance is formulated as a 7-tuple:
$$
(T, P, I, M, S, S', I').
$$
Here, $T$ denotes the sensitive privacy type, and $M$ specifies the corresponding private region in the source image $I$. The surrogate image $S$ is generated by replacing the sensitive region indicated by $M$ with synthesized surrogate content. Given an editing instruction $P$, we apply MLLM-based image editing to both the original image $I$ and the surrogate image $S$, obtaining the edited source image $I'$ and the edited surrogate image $S'$, respectively. The edited source image $I'$ serves as the ground-truth target for local recovery, while $S'$ represents the only edited visual result returned by the MLLM under privacy-preserving interaction.

\subsection{Dataset Construction}

\paragraph{Privacy Attribute Taxonomy}


\begin{figure}[t]
\centering
\captionsetup{skip=2pt}
\includegraphics[width=\linewidth, trim=40 60 40 59, clip]{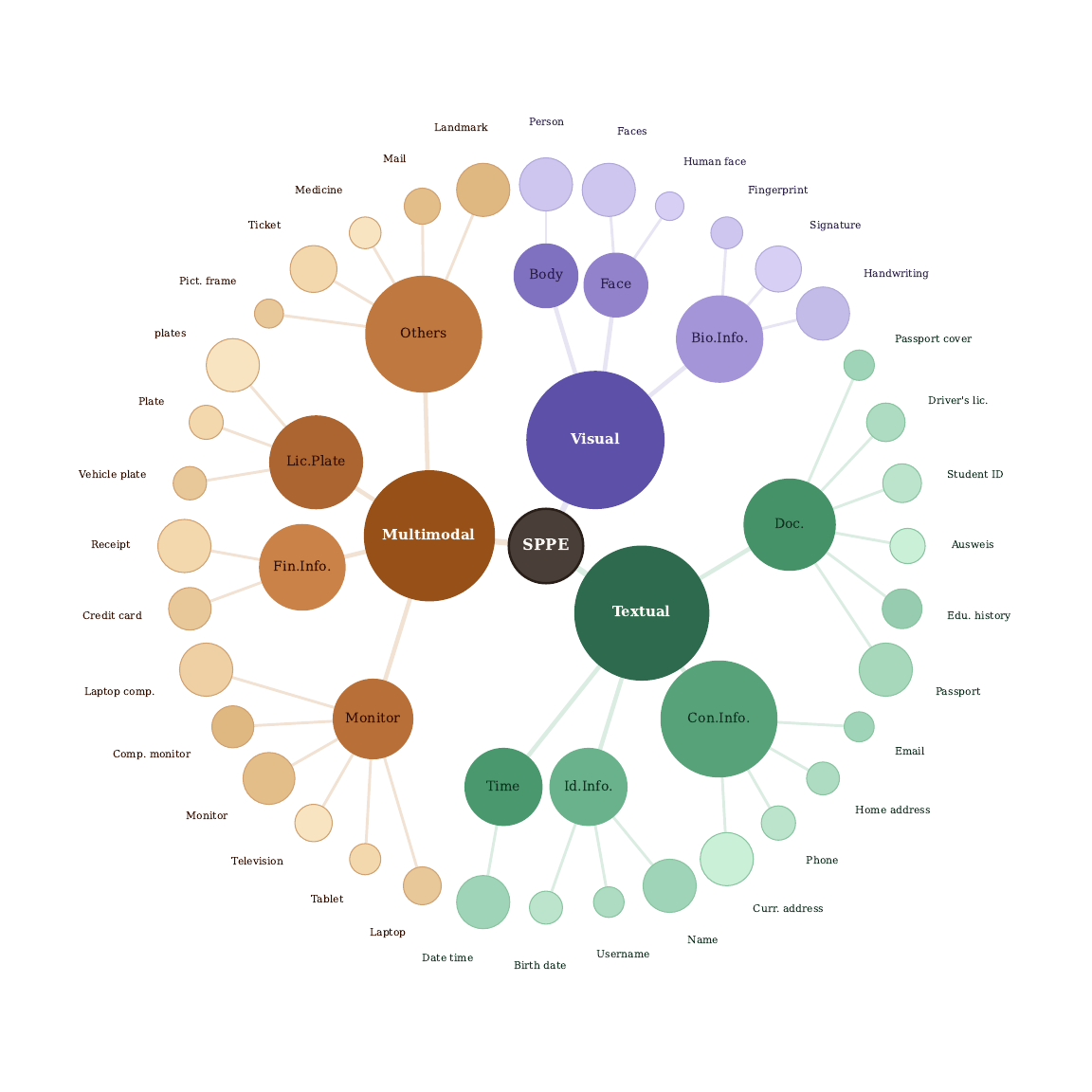}
\caption{Privacy attribute distribution in SPPE.}
\label{fig:data}
\vspace{-18pt}
\end{figure}
SPPE includes a wide range of privacy attributes organized into three modalities: visual, textual, and multimodal. These attributes are further grouped into 11 high-level privacy groups and 36 fine-grained privacy categories, as summarized in Table~\ref{tab:category_details}. The object-level privacy annotations are collected from established privacy-related image datasets, including VISPR~\cite{vispr}, Visual Redaction~\cite{visual_redaction}, DIPA~\cite{dipa}, and DIPA2~\cite{dipa2}. These source datasets contain human annotations and cover diverse privacy-sensitive scenarios, allowing the resulting taxonomy to reflect the practical diversity of what different individuals and communities consider private. Figure~\ref{fig:data} visualizes the category distribution, showing that SPPE covers common privacy-sensitive content such as faces, documents, contact information, financial information, and private environments.

\begin{table*}[t]
\centering
\footnotesize
\setlength{\tabcolsep}{3pt}
\renewcommand{\arraystretch}{0.92}
\captionsetup{skip=2pt}
\resizebox{\linewidth}{!}{
\begin{tabular}{c|l|p{0.58\linewidth}}
\toprule
\textbf{Privacy Modality} & \textbf{Privacy Group} & \textbf{Privacy Category (Count)} \\
\midrule

\multirow{3}{*}{Visual}
& Face &
human face (276), faces  (9366) \\

& Body &
person (434) \\

& Biometric Info &
fingerprint (92), signature (664), handwriting (1150) \\

\midrule
\multirow{5}{*}{Textual}
& Document &
passport cover (28), drivers license (130), student id (134),
ausweis (80), education history (152), passport (506) \\

& Contact Info &
email (136), home address (280), phone (372),
current address (2786) \\

& Identity Info &
name all (2508), birth date (260), username (146) \\

& Time &
date time (3020) \\

\midrule
\multirow{5}{*}{Multimodal}
& Monitor &
laptop (28), tablet computer (8), television (26),
monitor (108), computer monitor (46), laptop computer (118) \\

& Financial Info &
credit card (180), receipt (436) \\

& License Plate &
vehicle registration plate (96), plate (110),
multi-vehicle plates (858) \\

& Others &
picture frame (60), ticket (1074), medicine (136),
mail (314), landmark (1730) \\

\bottomrule
\end{tabular}
}
\caption{Detailed category breakdown of privacy attributes in the SPPE dataset.}
\label{tab:category_details}
\vspace{-10pt}
\end{table*}

\begin{figure*}
\centering
\includegraphics[width=\linewidth]{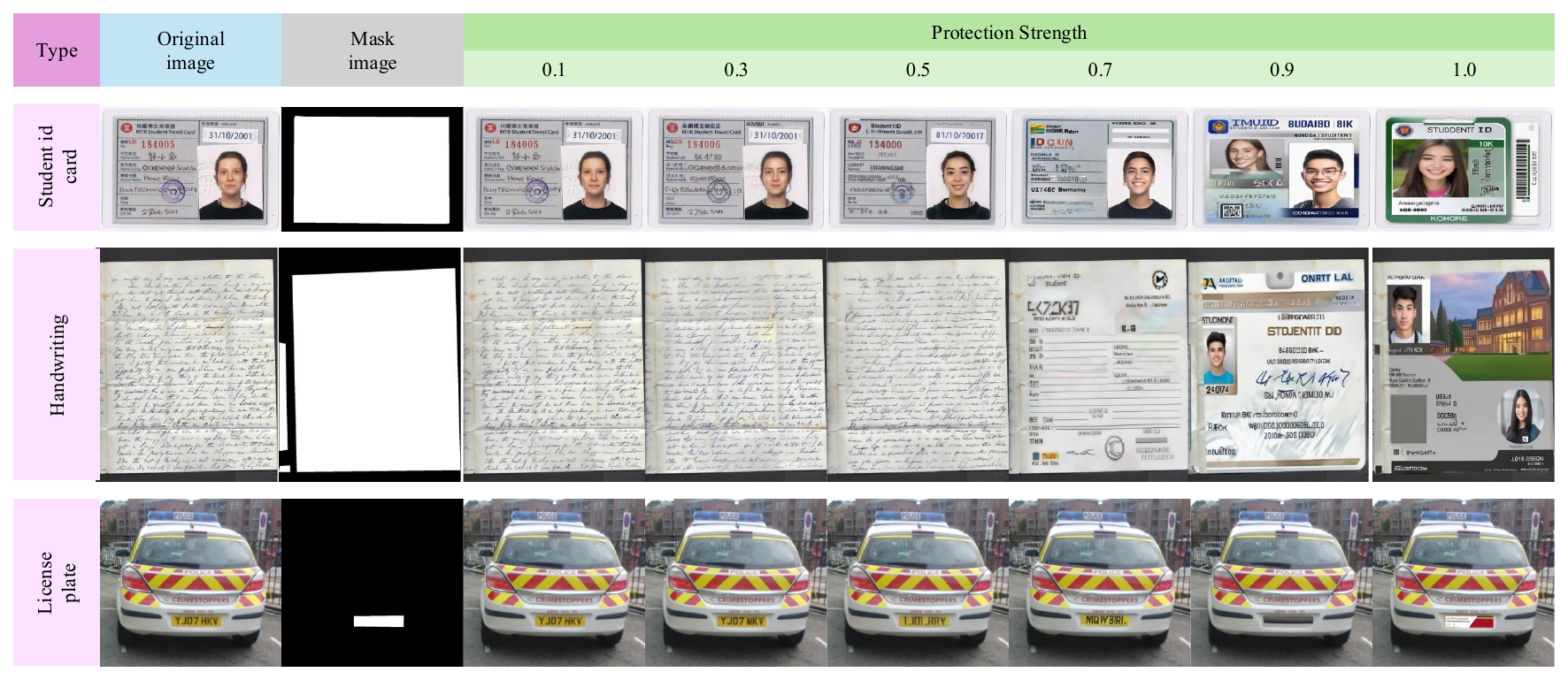}
\caption{Effect of guidance scale on surrogate generation. Increasing the guidance scale strengthens sensitive-region substitution but may also alter edit-relevant structure and context. SPPE adopts 0.5 as the default setting during benchmark construction.}
\label{fig:guidance}
\vspace{-10pt}
\end{figure*}

\paragraph{Surrogate Generation.} 

For each source image, SPPE creates a privacy-preserving surrogate by replacing the sensitive region with synthesized content using the SDXL inpainting model~\cite{sdxl}. This operation replaces directly exposed private information while preserving the surrounding layout and contextual cues needed by downstream image editing models. SPPE focuses on recovery utility under this surrogate-based setting, providing a standardized testbed for evaluating whether edited surrogate outputs can support source-domain recovery. During surrogate generation, the inpainting guidance scale affects how much the selected region is altered relative to its surrounding context. We compare representative guidance settings during dataset construction. As shown in Fig.~\ref{fig:guidance}, a very weak setting may retain more original visual traces, whereas a stronger setting may introduce larger changes to the protected region and its local semantics. We therefore use a guidance scale of 0.5 as the default setting to obtain surrogates with moderate regional substitution and contextual preservation.
This configuration is used as a standardized simulation of surrogate substitution for dataset construction.

\paragraph{Editing Instruction Design.} 
\begin{table*}
\centering
\footnotesize
\setlength{\tabcolsep}{3pt}
\renewcommand{\arraystretch}{0.92}
\captionsetup{skip=2pt}
\begin{tabular}{c|p{0.8\linewidth}}
  \toprule
\textbf{Type} & \textbf{Example Prompts} \\
\hline
Style & Turn to an oil painting style; Turn to a pencil sketch style... \\
Addition & Let the person wear sunglasses; Let the person wear a cowboy hat...\\
Removal & Remove the person; Remove the laptop; Remove the plate; Remove text... \\
Replace & Change the background to a night scene; Change the background to a sunset view... \\
Appearance & Change the person's hair color to red; Make the person look older... \\
Concept  &  Make it look futuristic; Make the person look like a ghost; ...\\
\bottomrule
\end{tabular}
\caption{Examples of editing prompts in SPPE, categorized by editing type.}
\label{tab:editing_types}
\end{table*}
To emulate flexible real-world user interactions, we define 65 editing instructions covering diverse modification types, including style transfer, object addition, object removal, replacement, appearance modification, and concept transformation. Table~\ref{tab:editing_types} provides representative prompts from each type. These instructions are designed to test whether a surrogate can preserve the semantics required by different editing behaviors, ranging from global style changes to localized  manipulations.


\paragraph{MLLM Editing and Ground Truth Generation.}
We employ SmartEdit~\cite{smartedit} to generate MLLM-style edits for both the original image $I$ and its surrogate $S$.
SmartEdit supports complex instruction-based editing through MLLM reasoning and serves as a controllable proxy for MLLM-style image editing. 
 For each privacy-protected image pair, two editing instructions are sampled to produce paired edited results $(S'_1, I'_1)$ and $(S'_2, I'_2)$. The edited surrogate $S'$ simulates the cloud-returned result under privacy-preserving interaction, while the edited source $I'$ is used as the reference target for evaluating local recovery. This yields paired evidence of instruction effects across the source and surrogate domains.

\subsection{Benchmark Tasks}
\label{sec:bench task}
SPPE is organized around the central question of whether surrogate-based protection can still support faithful editing recovery on the original private image. Before invoking an MLLM, the system needs to know whether the protected surrogate still contains sufficient edit-relevant evidence. After the surrogate has been edited, the system must recover the edited result locally in the source domain. 
Accordingly, SPPE defines two benchmark tasks that correspond to these two complementary stages of this process. Specifically, Task 1, \textbf{Editability Assessment}, evaluates whether a protected surrogate remains editable under a given instruction, while Task 2, \textbf{Surrogate-to-Source Edit Recovery}, evaluates whether the edit observed on the surrogate can be faithfully transferred back to the private source image. 


\paragraph{Task 1: Editability Assessment.}

Editability assessment studies whether the downstream editability of a privacy-preserving surrogate, measured  by editing-direction consistency, can be predicted before invoking the cloud MLLM. This task serves as a pre-recovery feasibility assessment: if the surrogate no longer supports the intended edit, the edited surrogate returned by the MLLM will provide weak or misleading guidance for local recovery. Thus, editability is not a generic measure of image quality or source-surrogate similarity. A surrogate may remain visually plausible while discarding the object, attribute, or contextual cue that determines the downstream edit. Conversely, a surrogate with noticeable appearance changes may still be useful if the prompt-relevant semantics are preserved. Given a source image $I$, its surrogate $S$, and an editing instruction $P$, The task predicts an editability score that estimates whether S can be effectively edited and provide reliable evidence for subsequent recovery.

\begin{figure}[t]
    \centering
    \includegraphics[width=\linewidth]{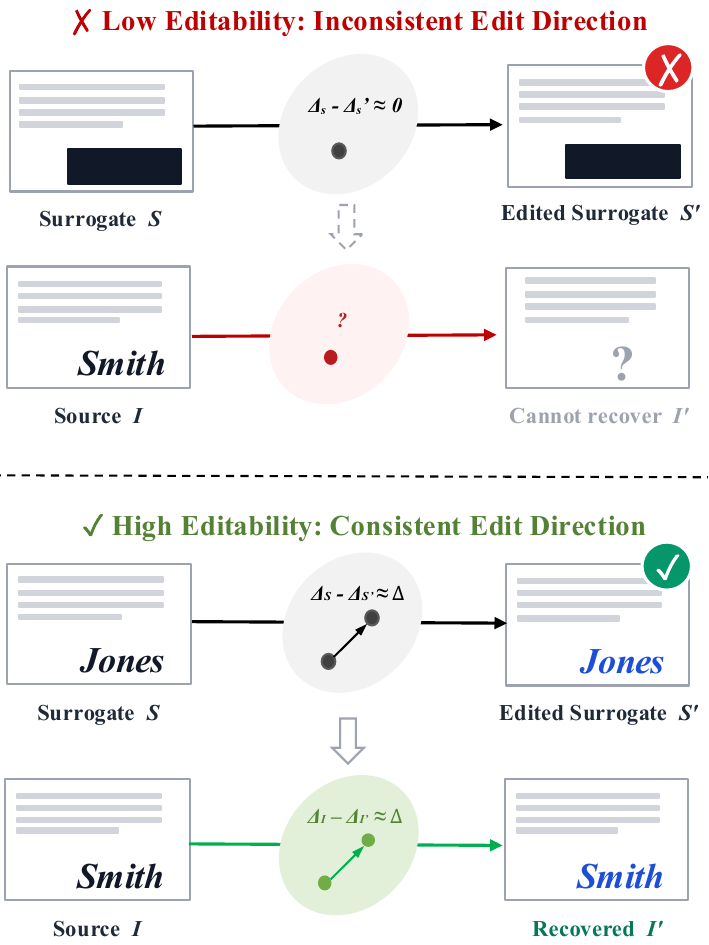}
    \caption{Editability assessment via semantic editing-direction consistency. The source-domain edit direction $(I \rightarrow I')$ is compared with the surrogate-domain edit direction $(S \rightarrow S')$ in the CLIP semantic space, and their alignment is used as a editability label for assessment.}
    \label{fig:task1}
\vspace{-14pt}
\end{figure}
SPPE operationalizes this notion through paired editing outcomes. As illustrated in Fig.~\ref{fig:task1}, an editable surrogate should induce a surrogate-domain edit $S'$ whose semantic change is consistent with the source-domain edit $I'$. This consistency reflects whether the surrogate preserves transferable edit semantics: if the edit induced on $S$ follows the same semantic direction as the edit induced on $I$, then the edited surrogate is more likely to provide useful evidence for local recovery. Inspired by~\cite{visii}, which measures editing performance by comparing the semantic direction from the source image to the edited image, we define the editability label as follows:
\par\smallskip
\noindent\makebox[\linewidth][c]{%
  $\displaystyle y = \cos\big(E_{\text{clip}}^{v}(I') - E_{\text{clip}}^{v}(I), \; E_{\text{clip}}^{v}(S') - E_{\text{clip}}^{v}(S)\big),$%
}
\par\smallskip
\noindent
where $E_{\text{clip}}^{v}(\cdot)$ denotes the CLIP image encoder. A higher score indicates that the surrogate preserves edit-relevant semantics more effectively and therefore has higher transferability for downstream recovery. Note that computing $y$ requires the edited images $I'$ and $S'$, which are generated offline during benchmark construction and used to derive editability labels for both training and test splits. The assessment model is trained to approximate this score from the pre-edit inputs $(I,S,P)$ alone, such that at inference time no edited images are required. We defer the metric-level evaluation protocol to the Experiment section~\ref{sec:experiment}.

\paragraph{Task 2: Surrogate-to-Source Edit Recovery.}
The second task evaluates the final utility of surrogate-based protection: recovering the edit performed on a surrogate back onto the original private image. This task directly measures the outcome that the overall framework is designed to optimize. Since the edited surrogate $S'$ is only an intermediate cloud-side output, it is insufficient for the protected interaction to merely obtain a plausible edited surrogate. A successful system must translate the edit effect expressed in $S'$ into the source domain while preserving private identity, fine-grained source details, and non-edited regions from $I$. In this sense, Task 2 subsumes the practical value of the surrogate: a surrogate is useful only insofar as it supports accurate local recovery after editing.

The recovery module operates on locally available inputs $(I, S, S', M, P)$ and produces a recovered edited image $\hat{I}$. SPPE evaluates $\hat{I}$ from two perspectives.
Source integrity measures its preservation of the original source $I$ across semantic, structural, and pixel-level dimensions. 
Edit consistency measures whether $\hat{I}$ follows the desired transformation, using both content similarity to the ground-truth edited source $I'$ and direction alignment between the trajectory $I \rightarrow \hat{I}$ and the reference trajectories $I \rightarrow I'$ or $S \rightarrow S'$  . Detailed metrics and protocols are specified in the Experiment section~\ref{sec:experiment}.

\begin{figure*}[t]
    \centering
    \includegraphics[width=1\linewidth]{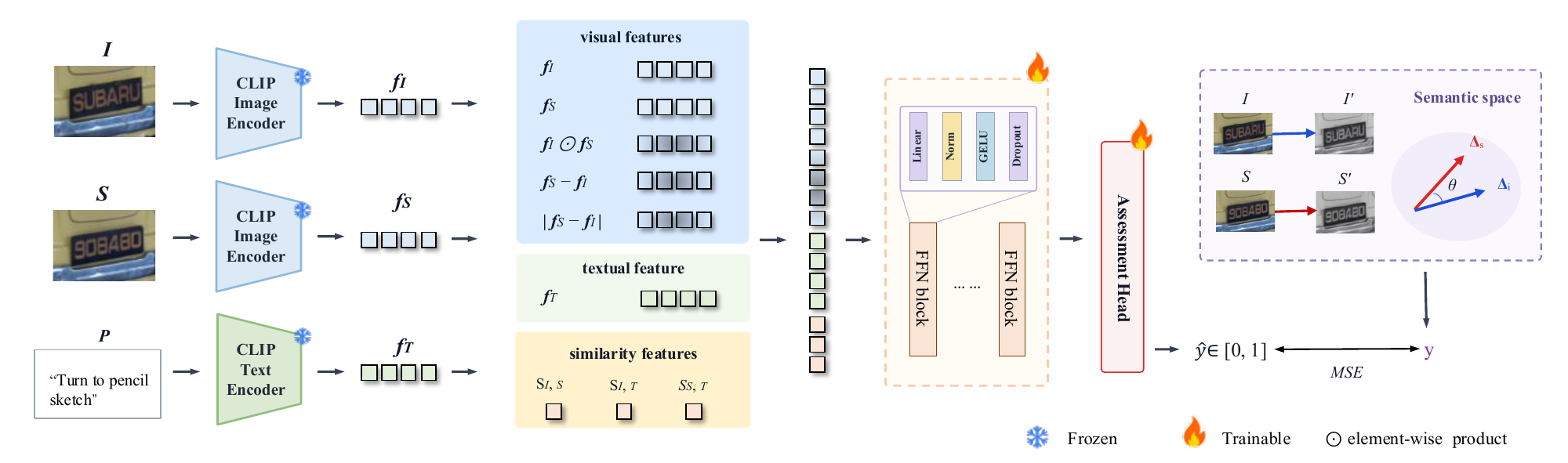}
\caption{
Architecture of ERMA for editability-aware surrogate assessment. 
Given the source image $I$, surrogate image $S$, and prompt $P$, frozen CLIP encoders extract visual features $f_I$, $f_S$ and textual feature $f_T$. 
ERMA constructs an editability-oriented relation representation using visual preservation cues. 
The fused representation is processed by a trainable assessment head to predict $\hat{y}$. 
}

    \label{fig:erma}
\end{figure*}
\section{ERMA: Editability-aware Relational Multi-modal \\ Assessment}

\subsection{Overview}
In surrogate-based privacy-preserving image editing, the role of a surrogate image is not merely to look similar to the original image, but to preserve the visual semantics required for subsequent instruction-guided editing. A surrogate with high perceptual similarity may still fail if it removes the object, attribute, or contextual cue required by the editing prompt. Conversely, a surrogate with visible appearance changes may remain effective as long as the edit-relevant semantics are preserved. Therefore, conventional image quality assessment metrics are insufficient for this setting, since they mainly measure perceptual distortion rather than downstream editability.

To address this issue, we propose \textbf{ERMA}, an Editability-aware Relational Multi-modal Assessment model. As shown in Fig.~\ref{fig:erma}, ERMA estimates editability from the source image, its privacy-preserving surrogate, and the editing prompt. Given a source image $I$, its surrogate $S$, and an editing prompt $P$, ERMA predicts an editability score:
\begin{equation}
    \hat{y} = \mathcal{A}_{\theta}(I,S,P),
\end{equation}
where a higher score indicates that the surrogate is more likely to induce editing behavior consistent with the original image. Different from standard full-reference IQA, ERMA explicitly models the relationship among the source image, surrogate image, and editing instruction. This allows the model to assess whether the surrogate preserves editable semantics rather than generic visual similarity alone.

\subsection{Relational Editability Modeling}

The central challenge of editability assessment lies in estimating whether the semantic transformation caused by surrogate generation will affect the intended edit. This requires a representation space where image content and language instruction can be compared in a unified manner. Therefore, ERMA adopts frozen CLIP encoders to project both images and prompts into a shared vision-language semantic space:
\begin{equation}
    f_I = E_{\text{clip}}^{v}(I), \quad
    f_S = E_{\text{clip}}^{v}(S), \quad
    f_T = E_{\text{clip}}^{t}(P),
\end{equation}
where $E_{\text{clip}}^{v}$ and $E_{\text{clip}}^{t}$ denote the CLIP image and text encoders, respectively. The CLIP encoders are kept frozen to provide stable semantic representations and avoid overfitting the assessment model to low-level dataset-specific distortions.

Using CLIP features is important for our task because editability is fundamentally semantic and instruction-dependent. For example, whether a surrogate is valid depends on whether it retains the content that the prompt intends to edit. A license plate surrogate may be visually different from the original, but it can still be editable if the structure and context required by a ``turn to pencil sketch'' instruction are preserved. In contrast, if the surrogate removes or distorts prompt-relevant semantics, the downstream MLLM edit may no longer correspond to the edit on the original image.

Based on this motivation, ERMA constructs a relational representation that captures both source-surrogate preservation and prompt-conditioned relevance. Specifically, we consider the source feature $f_I$, surrogate feature $f_S$, and text feature $f_T$ together with their relation cues:
\begin{equation}
\begin{aligned}
x = [&
f_I;\,
f_S;\,
f_I \odot f_S;\,
f_S - f_I;\,
|f_S - f_I|; \\
&
f_T;\,
s_{I,S};\,
s_{I,T};\,
s_{S,T}
].
\end{aligned}
\end{equation}
where $\odot$ denotes element-wise multiplication, and $s_{I,S}$, $s_{I,T}$, and $s_{S,T}$ are cosine similarities computed in the normalized CLIP feature space.

Each component is designed to reflect a different aspect of surrogate editability. The original and surrogate features provide the basic semantic content before and after privacy protection. The element-wise product captures their feature-level agreement, indicating how much semantic information is preserved. The signed difference $f_S-f_I$ describes the direction of the surrogate-induced semantic shift, while the absolute difference $|f_S-f_I|$ measures the magnitude of this deviation. These two discrepancy cues are useful because editability depends not only on how much the surrogate changes, but also on what kind of semantic shift is introduced.

The prompt-related similarities further make the assessment instruction-aware. The similarity $s_{I,T}$ reflects how strongly the original image is related to the editing instruction, while $s_{S,T}$ measures whether this prompt-relevant semantic information is still retained in the surrogate. The source-surrogate similarity $s_{I,S}$ provides a global semantic preservation cue. Together, these relation features allow ERMA to judge surrogate quality from the perspective of downstream editing behavior, rather than treating all visual changes as equally harmful.

\subsection{Editability Prediction and Optimization}

The relational representation $x$ is fed into a lightweight trainable assessment head to predict the editability score:
\begin{equation}
    \hat{y} = \sigma(H_{\theta}(x)),
\end{equation}
where $H_{\theta}$ denotes the trainable editability predictor and $\sigma(\cdot)$ constrains the output to $[0,1]$. With frozen CLIP encoders, ERMA trains only the predictor that maps the relational representation to the editability score.

ERMA is supervised by the editability label $y$ defined in Sec.~\ref{sec:bench task}. The label is computed offline from paired source-domain and surrogate-domain MLLM editing outcomes. Since the raw editing-direction consistency is measured by a cosine similarity $y_{\cos}\in[-1,1]$, we linearly map it to $y=(y_{\cos}+1)/2\in[0,1]$ for regression. ERMA predicts $\hat{y}$ from the pre-edit inputs $(I,S,P)$ and is optimized by:
\begin{equation}
    \mathcal{L}_{\mathrm{ERMA}}
    =
    \left\|
    \hat{y} - y
    \right\|_2^2 .
\end{equation}
At inference time, ERMA estimates surrogate editability from $(I,S,P)$ before editing.

\begin{figure*}[t]
    \centering
    \includegraphics[width=\linewidth]{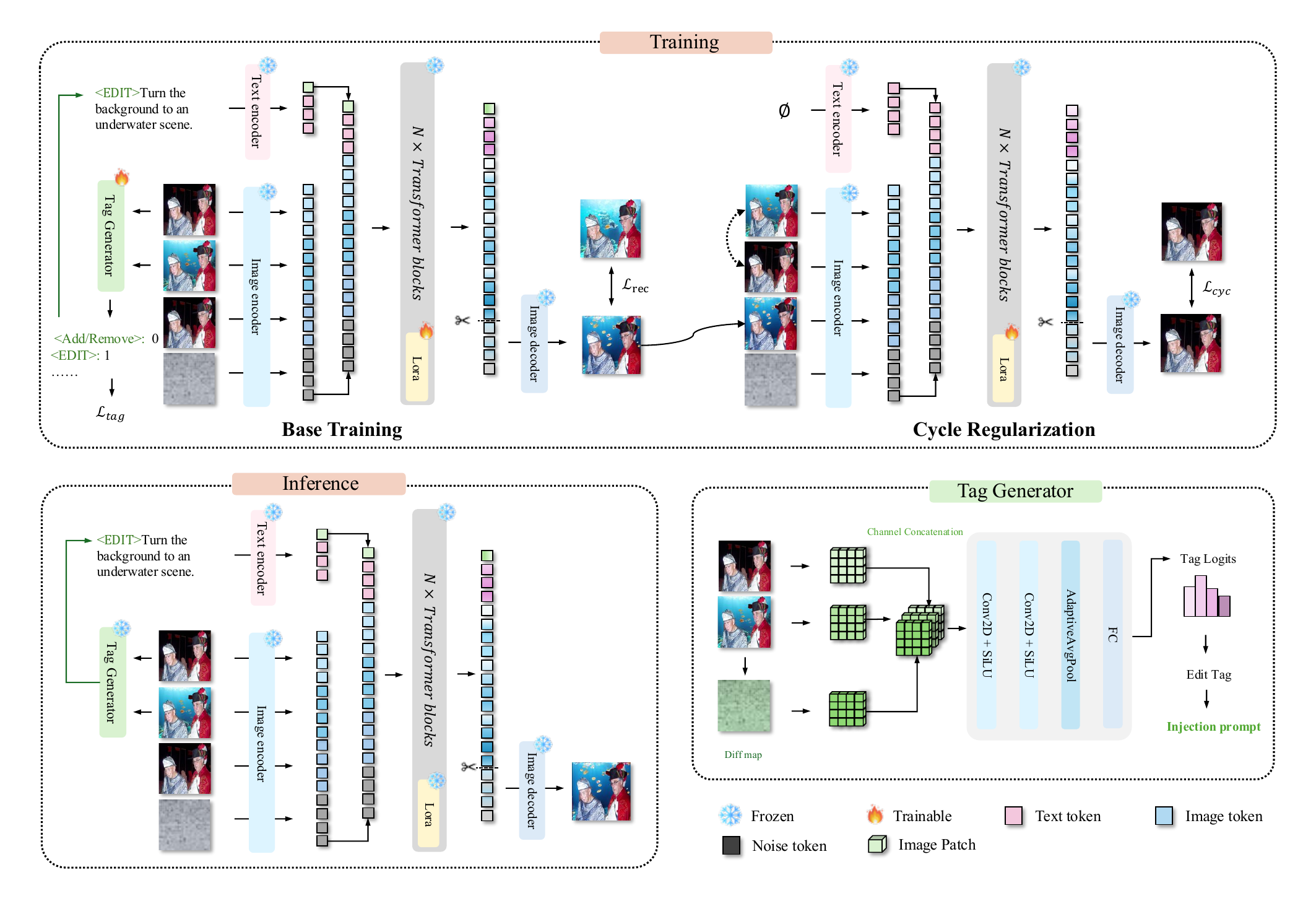}
    \caption{Overview of \method. The model uses the source image, surrogate pair, privacy mask, and editing instruction to recover the edited private image. The edit-conditioned tag summarizes the transformation observed in the surrogate domain, visual reference tokens provide source and edit evidence, and cycle-consistent regularization encourages accurate edit transfer while preserving source content.}
    \label{fig:s2ser}
\end{figure*}
\section{\method: \textbf{C}ycle-\textbf{C}onsistent \textbf{E}dit-Conditioned \\ \textbf{S}urrogate-to-\textbf{S}ource \textbf{E}dit \textbf{R}ecovery}

\subsection{Overview}

After the cloud-based MLLM edits a protected surrogate, the local system must recover the corresponding edited result on the original private image. This recovery problem is different from standard image editing: the model should not rely only on the text prompt, because the actual MLLM edit is reflected in the surrogate pair $(S,S')$, and it should not directly copy $S'$, because the surrogate no longer contains the original private content. To address this problem, we propose \textbf{\method}, a Cycle-Consistent Edit-Conditioned Surrogate-to-Source Edit Recovery model. As shown in Fig.~\ref{fig:s2ser}, \method jointly uses the source image $I$, surrogate image $S$, edited surrogate $S'$, mask $M$, and prompt $P$ to recover the edited source image:
\begin{equation}
\hat{I}=\mathcal{R}_{\theta}(I,S,S',M,P).
\end{equation}
Here, $I$ anchors source identity and private details, $(S,S')$ provides the observed edit evidence, $M$ marks the sensitive region requiring careful reconstruction, and $P$ describes the user's editing intent. The following modules explain how these inputs are converted into textual and visual conditions for recovery.

\subsection{Edit-Conditioned Tag Guidance}

The text prompt $P$ describes the user's intended edit, but it does not fully specify how the cloud MLLM realizes that edit on a particular image. For example, an instruction such as ``make the image look more modern'' may turn an old car into a newer model in one image, while in another image it may mainly change the overall color tone, materials, lighting, or background style. The surrogate editing pair $(S,S')$ therefore contains useful task evidence beyond the raw instruction, because it reveals the concrete transformation selected by the MLLM for the current image.

Inspired by prompt-based task conditioning~\cite{sanh2021multitask} and task-aware visual reasoning~\cite{xu2024fakeshield}, \method introduces an edit-conditioned tag generator to make this evidence explicit.The generator observes the surrogate before and after editing and predicts a coarse edit tag from three categories: $\langle ADD/REMOVE\rangle$ 
for object-level addition or removal operations, $\langle
EDIT\rangle$ for local attribute or appearance modifications, and $\langle STYLE\rangle$ for global style and semantic transformations. The intuition is that these tags provide a compact description of the dominant edit behavior, helping the recovery model decide whether it should transfer a global style, modify a local object or regenerate a region. Rather than using the tag as a hard rule, we convert it into a special token and prepend it to the original instruction:
\begin{equation}
C_T = E_{\text{text}}(c_g \Vert P),
\end{equation}
where $c_g$ denotes the generated edit-conditioned token and $E_{\text{text}}$ is the text encoder. This task-aware textual condition helps the recovery model disambiguate what kind of transformation should be transferred from $S'$ to $I$. In this way, the model is encouraged to recover the actual surrogate-domain edit, instead of over-relying on a generic interpretation of the text prompt.

The tag generator is trained with \textit{weak supervision} from coarse prompt categories. Specifically, we first group editing prompts into high-level edit types according to their instruction templates, and use these prompt-derived groups as weak labels. The purpose of these labels is to provide a coarse grouping signal that organizes edits by their dominant visual behavior. Since the generator predicts tags from the surrogate editing pair rather than from the prompt alone, it learns to group visually similar transformations across different prompt templates without over-relying on the textual instruction to determine edit type. The tag loss is formulated as
\begin{equation}
\mathcal{L}_{\mathrm{tag}}
=
-\sum_{k=1}^{K} t_k \log q_k,
\end{equation}
where $t_k$ denotes the weak label and $q_k$ is the predicted probability of the $k$-th edit tag.

\subsection{Visual Condition Construction}

Textual conditioning alone is insufficient for this task because the recovery model must preserve source-specific details while transferring an edit observed in another image domain. We therefore build visual conditions from three complementary inputs.

The source image $I$ provides the private content and visual identity that should remain local. This is especially important for privacy-sensitive regions such as faces, license plates, signatures, documents, and personal objects. The surrogate image $S$ acts as an aligned, privacy-preserving counterpart of $I$. It tells the model how the protected content is spatially and semantically related to the original scene. The edited surrogate $S'$ provides direct evidence of the cloud-side editing result, including color changes, texture changes, object insertion or removal, and global style transformations.

We encode all visual inputs with a  VAE encoder , obtaining latent representations $z_I = E_{\text{vae}}(I)$, $z_S = E_{\text{vae}}(S)$, and $z_{S'} = E_{\text{vae}}(S')$, and concatenate them as $C_V$:
\begin{equation}
C_V = [z_I \Vert z_S \Vert z_{S'}].
\end{equation}
This condition gives the diffusion transformer a compact in-context example: $S \rightarrow S'$ shows the edit to transfer, while $S \leftrightarrow I$ indicates how to map the edit back to the source domain.
During denoising, generated image tokens attend to these visual references, allowing the model to preserve source content  and transfer the observed edit.

\subsection{MM-DiT-based Recovery}

\method adopts a multi-modal diffusion transformer (MM-DiT) architecture as the generative backbone. This architecture is well suited to our recovery setting because the model must reason over heterogeneous conditions: text tokens describe the intended edit, visual reference tokens provide source and surrogate evidence, and noisy image tokens represent the image being generated. A transformer-based diffusion model can place these tokens in a shared attention space, allowing the generated image tokens to selectively attend to the prompt, the source image, and the surrogate editing pair during denoising.

Formally, we concatenate the textual condition, visual condition, and noisy image latent tokens into a unified sequence:
\begin{equation}
Z_t = [C_T \Vert C_V \Vert z_t].
\end{equation}
Through multi-modal self-attention, the recovery model updates the noisy image tokens by integrating semantic intent from $C_T$ and visual edit evidence from $C_V$. This design matches the structure of the recovery task: $S \rightarrow S'$ acts as an in-context edit demonstration, while $I$ provides the source-domain content to be edited.

During training, a latent state $z_t$ is sampled along the interpolation trajectory between Gaussian noise and the latent representation of the ground-truth edited source image $I'$. The model is optimized under the original flow-matching objective to recover the target edited image conditioned on the textual and visual inputs. During inference, the model starts from Gaussian noise and generates $\hat{I}$ under the same conditions. The prompt and generated tag guide the type of edit, the surrogate pair provides concrete edit evidence, and the source image anchors the recovered result to the private original.

\subsection{Cycle-Consistent Recovery Regularization}

Forward recovery alone supervises the model to produce an image close to $I'$, but it does not explicitly require the learned transformation to preserve the source content in an invertible or controllable way. This can lead to over-editing: the generated result may follow the surrogate edit but unnecessarily alter unrelated details.

To improve source integrity, we add a cycle-consistent regularization path. After the model predicts the edited source $\hat{I}$, we reverse the editing direction and ask the same model to reconstruct the original source image $I$ from the reversed evidence:
\begin{equation}
\tilde{I}=\mathcal{R}_{\theta}(\hat{I},S',S,M,\varnothing).
\end{equation}
Here, $\varnothing$ denotes an empty text condition. The model architecture accepts a text input at every forward pass. For the reverse path, we supply an empty condition because the transformation to be undone is fully specified by the visual pair $(S',S)$, and the reverse path is designed as a visual consistency regularizer rather than a semantic editing step. This lets the model focus on recovering $I$ from the observed visual correspondences in $(\hat{I},S',S)$, encouraging it to preserve source information that remains recoverable after the forward edit. To avoid noisy cycle supervision at early training stages when $\hat{I}$ may be unreliable, the cycle objective is introduced only after an initial forward recovery warm-up phase (see Training Setup in Section~\ref{sec:experiment}).

\subsection{Optimization and Parameter-efficient Fine-tuning}

The recovery model is trained with three complementary objectives. The forward objective teaches the model to recover the edited source image from surrogate-domain evidence, the cycle objective encourages source-preserving control, and the tag classification objective trains the weakly supervised edit-conditioned tag generator:
\begin{equation}
  \mathcal{L} = \mathcal{L}_{\text{rec}} +  \mathcal{L}_{\text{tag}} + \mathbf{1}(t > T_w) \mathcal{L}_{\text{cycle}}
\end{equation}
Here, $\mathcal{L}_{\text{rec}}$ and $\mathcal{L}_{\text{cycle}}$ denote the flow-matching loss. We adopt a staged optimization schedule with a warm-up boundary $T_w$. During the warm-up stage ($t \leq T_w$), the model is optimized only with the forward recovery objective so that it can first learn a stable mapping. After the warm-up stage ($t > T_w$), the cycle-consistent recovery regularizer is introduced for continued fine-tuning. This design avoids enforcing the reverse constraint before the forward recovery behavior becomes stable.

To adapt the pretrained diffusion transformer without destroying its generative prior, we fine-tune the model with Low-Rank Adaptation (LoRA). Surrogate-to-source recovery requires the model to learn a new conditional behavior from limited paired data: it must interpret $(S,S')$ as edit evidence and apply the corresponding change to $I$. Fully fine-tuning the backbone may overfit to SPPE and weaken the model's general image prior, while freezing all parameters would limit adaptation to this new input structure. LoRA offers a balanced solution by injecting trainable low-rank updates into attention layers, allowing the model to learn task-specific cross-modal interactions among text, source, surrogate, and noisy image tokens. Since only lightweight low-rank parameters are updated while the backbone remains frozen, training is more efficient and the pretrained generation capability is better preserved.

\section{Experiment}
\label{sec:experiment}
\noindent \textbf{Dataset.}
We evaluate our method on both the proposed \textbf{SPPE} dataset and the public \textbf{InstructPix2Pix} dataset. For fair comparison, all trainable methods are trained on the SPPE training split. Evaluation is conducted on the SPPE test set and a subset of the InstructPix2Pix dataset, covering 934 unique prompts. Since InstructPix2Pix lacks sensitive region annotations, we generate full-image surrogates for its samples. While InstructPix2Pix contains synthetic images and prompts absent from SPPE, it offers a comprehensive benchmark for assessing model generalization to real-world scenarios where users may provide novel edit instructions.

\noindent \textbf{Baselines.}
\textit{Task 1: Editability Assessment.}
We compare ERMA with representative full-reference image quality assessment and perceptual similarity methods. These baselines estimate the quality or similarity between a source image and its surrogate, providing a natural test of whether generic visual preservation is sufficient for predicting editability. AFINE-FR~\cite{AFINE-FR} studies generalized full-reference IQA by relaxing the assumption that the reference image is perfectly pristine. AHIQ~\cite{ahiq} combines CNN and transformer-style attention features for hybrid image quality assessment. DISTS~\cite{dists} measures deep structure and texture similarity, making it suitable for evaluating whether a surrogate preserves perceptual content. ST-LPIPS~\cite{STLPIPS} extends learned perceptual similarity with shift-tolerant design, reducing sensitivity to small spatial misalignments. TOPIQ-FR~\cite{TOPIQ} adopts a top-down IQA strategy that uses high-level semantic cues to guide distortion assessment in important regions. Since these methods are not instruction-aware, they are evaluated as source-surrogate assessment baselines, while ERMA additionally conditions on the editing prompt.

\textit{Task 2: Surrogate-to-Source Edit Recovery.}
We compare \method with representative methods related to visual instruction inversion, in-context image editing, reference-based edit transfer, and surrogate-driven recovery. VISII~\cite{visii} performs image editing via visual instruction inversion, using image-prompting cues to infer the desired editing direction (note that VISII uses InstructPix2Pix data during training and is therefore excluded from the InstructPix2Pix generalization evaluation). Prompt-Diffusion~\cite{prompt-diffusion} is an in-context diffusion baseline that learns to perform image editing from visual examples, making it relevant to our setting where $(S,S')$ provides an edit demonstration. EditTransfer~\cite{edittransfer} learns image editing through vision in-context relations and transfers the observed edit from a reference pair to a target image. Cross-Image Attention~\cite{cross_image} is a zero-shot appearance transfer method that uses cross-image attention to propagate visual attributes from a reference image, serving as a reference-based transfer baseline. We also include SOER~\cite{aaai26}, a surrogate-oriented edit recovery method designed for privacy-preserving MLLM editing. For methods not originally formulated for surrogate-to-source recovery, we adapt their inputs to use the source image and surrogate editing pair under the same SPPE training and evaluation protocol.

All trainable methods are trained on the SPPE training split for fair comparison.

\noindent \textbf{Evaluation Metrics.}
\textit{Task 1: Editability Assessment.}
We evaluate how well each method predicts the editability scores defined in SPPE. Following common quality assessment protocols, we report Spearman's Rank Correlation Coefficient (SRCC) and Pearson's Linear Correlation Coefficient (PLCC) between predicted scores and editability scores. SRCC measures rank-order consistency, indicating whether a method can correctly order surrogates from less editable to more editable. PLCC measures linear prediction accuracy after score calibration, reflecting how closely the predicted scores match the continuous editability scores. Together, these two metrics evaluate both monotonic agreement and score-level fidelity.

\textit{Task 2: Surrogate-to-Source Edit Recovery.}
We evaluate two complementary objectives: \textit{Edit Consistency} and \textit{Source Integrity}. Edit Consistency measures whether the recovered image $\hat{I}$ reflects the target editing effect. We compare $\hat{I}$ with the ground-truth edited source image $I'$ using CLIP Similarity (CLIP-Sim), Structural Similarity Index (SSIM), and Peak Signal-to-Noise Ratio (PSNR). We also report directional similarity metrics~\cite{visii, instanip}: DirI compares the editing directions $(I \rightarrow \hat{I})$ and $(I \rightarrow I')$, while DirS compares $(I \rightarrow \hat{I})$ and $(S \rightarrow S')$, measuring whether the recovered result follows the source-domain and surrogate-domain edit trajectories.
Source Integrity measures whether $\hat{I}$ preserves the original source content, especially regions unrelated to the intended edit. We compute CLIP-Sim, SSIM, and PSNR between $\hat{I}$ and $I$ as source integrity metrics.
Because high source integrity scores alone may reflect under-editing, we jointly consider source integrity and edit consistency to assess whether a method preserves the source while realizing the intended edit.

\noindent \textbf{Training Setup.}
Our method builds on the FLUX Inpainting model, which adopts a DiT-based architecture with a T5 text encoder and a VAE image encoder. We fine-tune the model on the SPPE training split using LoRA (rank 256) for 6000 steps: the first 3000 steps use forward recovery warm-up with edit-conditioned tag supervision, and the remaining steps additionally introduce the cycle-consistent recovery regularizer. Optimization uses 8-bit AdamW with a learning rate of $1 \times 10^{-4}$. 

\subsection{Task 1: Editability Assessment}
\subsubsection{Quantitative Results}

\begin{table*}[t]
\centering
\setlength{\tabcolsep}{2.5pt}
\resizebox{\linewidth}{!}{%
\begin{tabular}{lcccccc}
\toprule
\textbf{Metric} & \textbf{AFINE-FR~\cite{AFINE-FR}} & \textbf{AHIQ~\cite{ahiq}} & \textbf{DISTS~\cite{dists}} & \textbf{ST-LPIPS~\cite{STLPIPS}} &
\textbf{TOPIQ-FR~\cite{TOPIQ}} & \textbf{ERMA} \\
\midrule
\textbf{SRCC} & 0.5345 & 0.3963 & \latexunderline{0.5843} & 0.5519 & 0.5302 & \textbf{0.6655} \\
\textbf{PLCC} & 0.4404 & 0.4733 & 0.5917 & 0.4595 & \latexunderline{0.6079} & \textbf{0.6826} \\
\bottomrule
\end{tabular}
}
\caption{Task 1 editability assessment on SPPE. SRCC and PLCC measure ranking consistency and score fidelity against editability scores. \textbf{Bold} and \underline{underline}
 denote the best and second-best results.}
\label{tab:comparison}
\end{table*}
Table~\ref{tab:comparison} evaluates editability prediction from two perspectives: ranking consistency (SRCC), which measures whether a method can correctly order surrogates by editability, and score fidelity (PLCC), which measures how well the predicted scores fit the continuous editability scores. ERMA achieves the best result on both metrics, obtaining 0.6655 SRCC and 0.6826 PLCC. Compared with the strongest generic baselines, ERMA improves SRCC over DISTS from 0.5843 to 0.6655 (13.9\%) and PLCC over TOPIQ-FR from 0.6079 to 0.6826 (12.3\%). These gains indicate that instruction-aware relation modeling better captures edit-relevant semantics than image quality metrics that do not condition on the editing instruction.

\begin{figure*}[t]
\centering
\includegraphics[width=\linewidth]{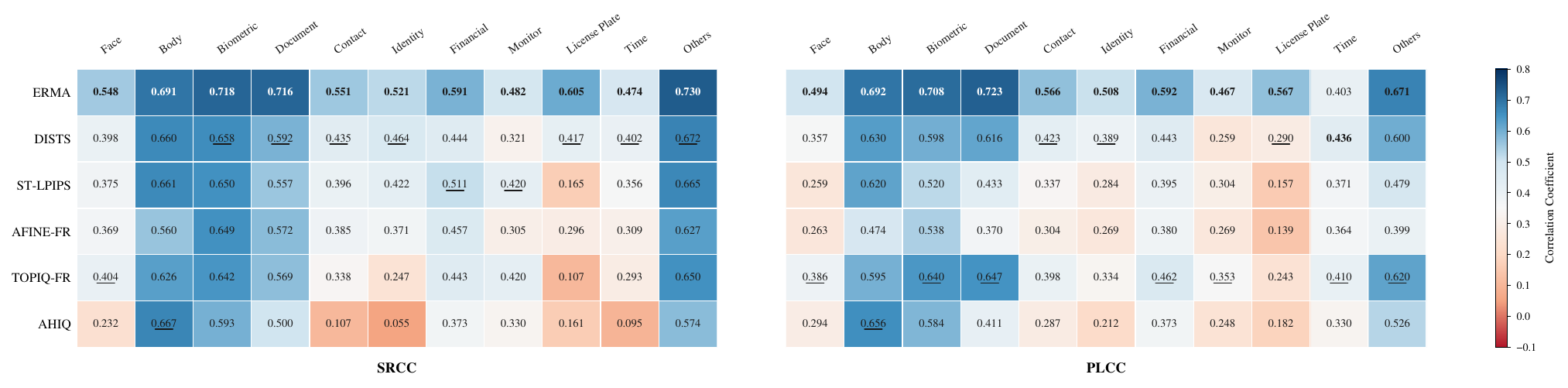}
\caption{Category-level SRCC and PLCC comparison for Task 1 editability assessment. Larger values indicate stronger agreement with editability scores.}
\label{fig:task1_heatmaps}
\end{figure*}

Fig.~\ref{fig:task1_heatmaps} further breaks down the results by privacy category and correlation type. The SRCC heatmap reflects whether a method preserves the correct ordering of surrogate editability within each category, while the PLCC heatmap reflects whether its predicted scores are numerically calibrated to the editability scores. ERMA shows consistently stronger correlations across both views, whereas full-reference IQA baselines vary substantially depending on the privacy type. The license plate and identity categories illustrate this gap most clearly: on license plates, ERMA achieves an SRCC of 0.605 and PLCC of 0.567, while DISTS falls to 0.417 and TOPIQ-FR to 0.107 in SRCC, indicating that perceptual similarity measures fail to capture whether surrogate changes in structured compact regions affect downstream editability. ERMA achieves the most consistent performance across categories (SRCC std = 0.092), compared to TOPIQ-FR (0.170) and AHIQ (0.211), suggesting that instruction-aware relation modeling generalizes more robustly across object types. Since editability is driven by the interaction between image content and editing instruction rather than the privacy label itself, cross-category stability is a practically important property.


\begin{table*}[t]
\caption{Comparison of methods across categories on source integrity metrics (CLIP-Sim (Src), SSIM (Src), PSNR (Src)), which measure how well the recovered image preserves the original private source. The proposed method is highlighted. \textbf{Bold} = best, \underline{underline} = second best.}
\label{tab:all_metrics}
\centering
\noindent\textbf{CLIP-Sim (Src)}\\
\resizebox{\textwidth}{!}{%
\begin{tabular}{l|ccccccccccc|c}
  \toprule
  \textbf{Method} & \textbf{Face} & \textbf{Body} & \textbf{Biometric} & \textbf{Document} & \textbf{Contact} & \textbf{Identity} & \textbf{Financial} & \textbf{Monitor} & \textbf{License Plate} & \textbf{Time} & \textbf{Others} & \textbf{Overall} \\
  \hline
    VISII~\cite{visii} & 0.6800 & 0.7662 & 0.6766 & \underline{0.7261} & \textbf{0.6851} & \textbf{0.6730} & \textbf{0.6933} & \textbf{0.8179} & 0.7407 & \textbf{0.6654} & \textbf{0.6727} & 0.6843 \\
    EditTransfer~\cite{edittransfer} & 0.7077 & 0.7618 & 0.6375 & 0.6527 & 0.6182 & 0.6021 & 0.5537 & 0.7743 & 0.7318 & 0.5872 & 0.5914 & 0.6491 \\
    Prompt-Diffusion~\cite{prompt-diffusion} & 0.6363 & 0.6636 & 0.5382 & 0.5508 & 0.5186 & 0.4964 & 0.5059 & 0.6972 & 0.6286 & 0.4880 & 0.5047 & 0.5610 \\
    Cross~\cite{cross_image} & 0.7277 & 0.7410 & 0.6444 & 0.6495 & 0.6302 & 0.6137 & 0.5761 & 0.7964 & 0.7377 & 0.5996 & 0.5936 & 0.6613 \\
    SOER~\cite{aaai26} & \underline{0.7381} & \underline{0.7979} & \underline{0.6913} & 0.7172 & \underline{0.6541} & 0.6362 & 0.6030 & 0.8154 & \underline{0.7462} & 0.6207 & 0.6366 & \underline{0.6856} \\
    \rowcolor{blue!10}\methodtable & \textbf{0.7472} & \textbf{0.8123} & \textbf{0.7084} & \textbf{0.7569} & 0.6531 & \underline{0.6381} & \underline{0.6547} & \underline{0.8167} & \textbf{0.7481} & \underline{0.6213} & \underline{0.6521} & \textbf{0.6946} \\
  \bottomrule
\end{tabular}
}

\noindent\textbf{SSIM (Src)}\\
\resizebox{\textwidth}{!}{%
\begin{tabular}{l|ccccccccccc|c}
  \toprule
  \textbf{Method} & \textbf{Face} & \textbf{Body} & \textbf{Biometric} & \textbf{Document} & \textbf{Contact} & \textbf{Identity} & \textbf{Financial} & \textbf{Monitor} & \textbf{License Plate} & \textbf{Time} & \textbf{Others} & \textbf{Overall} \\
  \hline
    VISII~\cite{visii} & 0.5932 & 0.6798 & 0.6660 & \textbf{0.6905} & \textbf{0.6920} & \textbf{0.6966} & \textbf{0.7100} & \textbf{0.7076} & \underline{0.6527} & \textbf{0.6780} & \textbf{0.6913} & \underline{0.6548} \\
    EditTransfer~\cite{edittransfer} & 0.6093 & 0.6162 & 0.6127 & 0.5689 & 0.6031 & 0.6080 & 0.5613 & 0.6067 & 0.6077 & 0.5771 & 0.5874 & 0.5998 \\
    Prompt-Diffusion~\cite{prompt-diffusion} & 0.4736 & 0.4940 & 0.3657 & 0.3853 & 0.3953 & 0.3707 & 0.3561 & 0.4289 & 0.4222 & 0.3508 & 0.3762 & 0.4113 \\
    Cross~\cite{cross_image} & 0.6358 & 0.6250 & 0.6359 & 0.5949 & 0.6294 & 0.6290 & 0.5977 & 0.6369 & 0.6264 & 0.6022 & 0.6139 & 0.6250 \\
    SOER~\cite{aaai26} & \underline{0.6530} & \underline{0.6833} & \underline{0.6661} & 0.6497 & 0.6438 & 0.6457 & 0.6101 & 0.6796 & 0.6408 & 0.6125 & 0.6374 & 0.6447 \\
    \rowcolor{blue!10}\methodtable & \textbf{0.6744} & \textbf{0.7047} & \textbf{0.6831} & \underline{0.6831} & \underline{0.6536} & \underline{0.6544} & \underline{0.6529} & \underline{0.7041} & \textbf{0.6546} & \underline{0.6219} & \underline{0.6585} & \textbf{0.6621} \\
  \bottomrule
\end{tabular}
}

\noindent\textbf{PSNR (Src)}\\
\resizebox{\textwidth}{!}{%
\begin{tabular}{l|ccccccccccc|c}
  \toprule
  \textbf{Method} & \textbf{Face} & \textbf{Body} & \textbf{Biometric} & \textbf{Document} & \textbf{Contact} & \textbf{Identity} & \textbf{Financial} & \textbf{Monitor} & \textbf{License Plate} & \textbf{Time} & \textbf{Others} & \textbf{Overall} \\
  \hline
    VISII~\cite{visii} & 16.0149 & \underline{17.7632} & \underline{17.5549} & \underline{17.9370} & \textbf{17.5727} & \underline{17.2665} & \textbf{17.7971} & \textbf{17.6609} & \textbf{16.3822} & \textbf{17.1885} & \textbf{17.5282} & \underline{16.9467} \\
    EditTransfer~\cite{edittransfer} & 14.8964 & 15.2879 & 16.4679 & 15.2926 & 15.2143 & 15.4483 & 14.5568 & 14.4330 & 14.4439 & 15.0307 & 15.0344 & 15.1299 \\
    Prompt-Diffusion~\cite{prompt-diffusion} & 12.4018 & 13.0876 & 11.2062 & 11.6294 & 11.6111 & 11.2937 & 11.4873 & 11.5592 & 11.3829 & 11.2314 & 11.2378 & 11.7413 \\
    Cross~\cite{cross_image} & 15.7560 & 15.7513 & 16.0118 & 15.4319 & 15.5150 & 15.4688 & 15.2183 & 15.5621 & 15.2017 & 15.1708 & 15.3239 & 15.5464 \\
    SOER~\cite{aaai26} & \underline{16.2601} & 17.3275 & 17.4731 & 17.0727 & 16.5880 & 16.5276 & 15.8928 & 15.8414 & 15.4774 & 15.8673 & 16.3142 & 16.3847 \\
    \rowcolor{blue!10}\methodtable & \textbf{17.2368} & \textbf{18.0678} & \textbf{18.2086} & \textbf{18.2102} & \underline{17.2953} & \textbf{17.3029} & \underline{17.4416} & \underline{17.3476} & \underline{16.2096} & \underline{16.5612} & \underline{17.2877} & \textbf{17.2635} \\
  \bottomrule
\end{tabular}
}

\end{table*}

\subsection{Task 2: Surrogate-to-Source Edit Recovery}

\subsubsection{Quantitative Results on SPPE}

Tables~\ref{tab:all_metrics} and~\ref{tab:all_metrics_edit} report recovery performance on the SPPE test set. \method achieves the best overall result across all eight evaluation metrics, with a 5.4\% improvement in PSNR (Src) and a 2.9\% improvement in PSNR (GT) over SOER. The gains extend to directional metrics, with \method leading on both DirI and DirS. At the overall aggregate level, \method leads on all eight metrics, with particularly strong category-level gains on biometric (2.5\% in CLIP-Sim (Src), 3.7\% in PSNR (GT)), contact (4.5\% in PSNR (GT)), and document (3.8\% in PSNR (GT)) over SOER. A notable comparison is VISII, which achieves competitive source integrity scores in some categories, primarily because it makes minimal visual changes and stays close to the input. This approach, however, yields substantially lower edit consistency: \method obtains DirI of 0.7411 compared to 0.5796 for VISII, confirming that source integrity and edit consistency must be read jointly.
\begin{figure*}
    \centering
    \captionsetup{skip=2pt}
    \includegraphics[width=\linewidth]{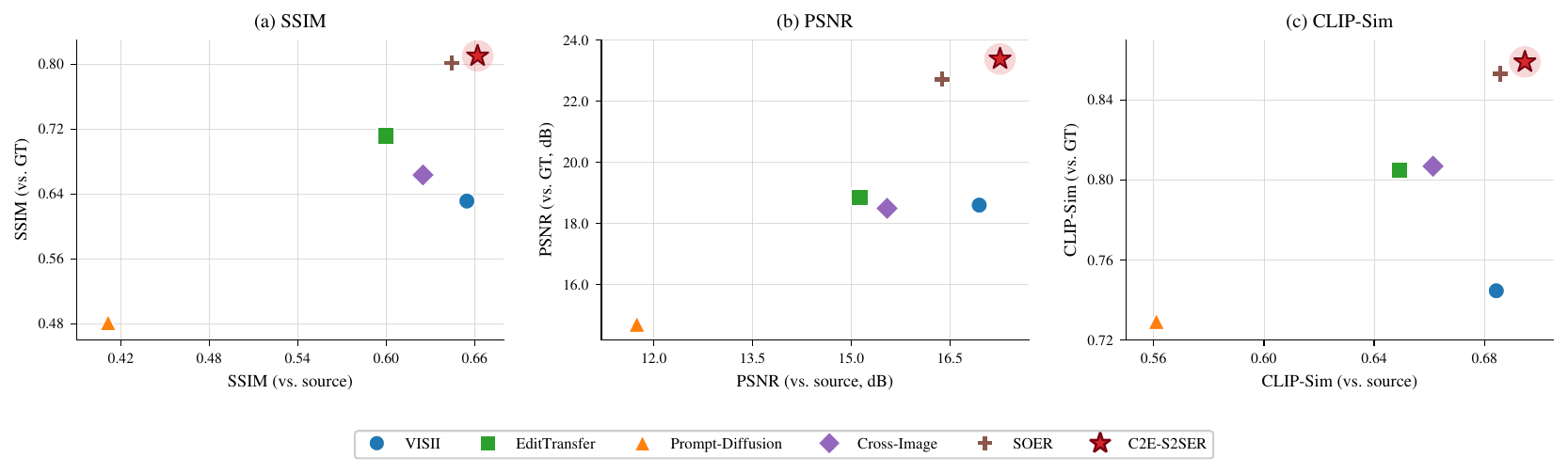}
\caption{Trade-off between source preservation and edit consistency on SPPE. The horizontal axis reports source-reference quality, while the vertical axis reports ground-truth-reference quality. Points closer to the upper-right indicate a better balance between preserving the private source and realizing the intended edit.}
\label{fig:tradeoff_recovery}
\end{figure*}

\begin{table*}[t]
\small
\caption{Comparison of methods across categories on edit consistency metrics (DirI, DirS, CLIP-Sim (GT), SSIM (GT), PSNR (GT)), which measure how faithfully the recovered image reflects the intended editing effect. The proposed method is highlighted. \textbf{Bold} = best, \underline{underline} = second best.}
\label{tab:all_metrics_edit}
\centering
\noindent\textbf{DirI}\\
\resizebox{\textwidth}{!}{%
\begin{tabular}{l|ccccccccccc|c}
  \toprule
  \textbf{Method} & \textbf{Face} & \textbf{Body} & \textbf{Biometric} & \textbf{Document} & \textbf{Contact} & \textbf{Identity} & \textbf{Financial} & \textbf{Monitor} & \textbf{License Plate} & \textbf{Time} & \textbf{Others} & \textbf{Overall} \\
  \hline
    VISII~\cite{visii} & 0.5036 & 0.5789 & 0.6293 & 0.5623 & 0.6305 & 0.6318 & 0.5969 & 0.5391 & 0.6208 & 0.6304 & 0.6029 & 0.5796 \\
    EditTransfer~\cite{edittransfer} & 0.6655 & 0.6287 & 0.6639 & 0.5767 & 0.7155 & 0.7150 & 0.6556 & 0.6038 & 0.7166 & 0.7272 & 0.6643 & 0.6817 \\
    Prompt-Diffusion~\cite{prompt-diffusion} & 0.6153 & 0.5832 & 0.5992 & 0.5500 & 0.6316 & 0.6295 & 0.6384 & 0.5494 & 0.6092 & 0.6434 & 0.6173 & 0.6181 \\
    Cross~\cite{cross_image} & 0.6452 & 0.6110 & 0.6650 & 0.5915 & 0.6968 & 0.7036 & 0.6914 & 0.6039 & 0.6815 & 0.7203 & 0.6687 & 0.6706 \\
    SOER~\cite{aaai26} & \textbf{0.7330} & \textbf{0.6918} & \underline{0.7083} & \textbf{0.6465} & \underline{0.7687} & \underline{0.7660} & \textbf{0.7147} & \textbf{0.6890} & \underline{0.7668} & \underline{0.7747} & \textbf{0.7076} & \underline{0.7382} \\
    \rowcolor{blue!10}\methodtable & \underline{0.7324} & \underline{0.6788} & \textbf{0.7101} & \underline{0.6343} & \textbf{0.7777} & \textbf{0.7748} & \underline{0.7065} & \underline{0.6845} & \textbf{0.7770} & \textbf{0.7866} & \underline{0.7069} & \textbf{0.7411} \\
  \bottomrule
\end{tabular}
}

\noindent\textbf{DirS}\\
\resizebox{\textwidth}{!}{%
\begin{tabular}{l|ccccccccccc|c}
  \toprule
  \textbf{Method} & \textbf{Face} & \textbf{Body} & \textbf{Biometric} & \textbf{Document} & \textbf{Contact} & \textbf{Identity} & \textbf{Financial} & \textbf{Monitor} & \textbf{License Plate} & \textbf{Time} & \textbf{Others} & \textbf{Overall} \\
  \hline
    VISII~\cite{visii} & 0.4340 & 0.3997 & 0.4544 & 0.3197 & 0.5621 & 0.5555 & 0.2489 & 0.4088 & 0.5668 & 0.5817 & 0.4020 & 0.4753 \\
    EditTransfer~\cite{edittransfer} & 0.6142 & 0.4717 & 0.5382 & 0.3486 & 0.6771 & 0.6713 & 0.3397 & 0.5417 & 0.6932 & 0.7050 & 0.4781 & 0.6027 \\
    Prompt-Diffusion~\cite{prompt-diffusion} & 0.5548 & 0.4328 & 0.4690 & 0.3329 & 0.5769 & 0.5702 & 0.3334 & 0.4680 & 0.5774 & 0.6045 & 0.4353 & 0.5306 \\
    Cross~\cite{cross_image} & 0.5862 & 0.4567 & 0.5460 & \underline{0.3719} & 0.6466 & 0.6466 & \underline{0.3810} & 0.5077 & 0.6484 & 0.6913 & 0.4853 & 0.5858 \\
    SOER~\cite{aaai26} & \textbf{0.6874} & \textbf{0.5316} & \underline{0.5909} & \textbf{0.3911} & \underline{0.7407} & \underline{0.7356} & \textbf{0.3888} & \textbf{0.6292} & \underline{0.7579} & \underline{0.7647} & \underline{0.5268} & \underline{0.6658} \\
    \rowcolor{blue!10}\methodtable & \underline{0.6873} & \underline{0.5037} & \textbf{0.5963} & 0.3704 & \textbf{0.7586} & \textbf{0.7501} & 0.3649 & \underline{0.6189} & \textbf{0.7769} & \textbf{0.7867} & \textbf{0.5296} & \textbf{0.6719} \\
  \bottomrule
\end{tabular}
}

\noindent\textbf{CLIP-Sim (GT)}\\
\resizebox{\textwidth}{!}{%
\begin{tabular}{l|ccccccccccc|c}
  \toprule
  \textbf{Method} & \textbf{Face} & \textbf{Body} & \textbf{Biometric} & \textbf{Document} & \textbf{Contact} & \textbf{Identity} & \textbf{Financial} & \textbf{Monitor} & \textbf{License Plate} & \textbf{Time} & \textbf{Others} & \textbf{Overall} \\
  \hline
    VISII~\cite{visii} & 0.7157 & 0.8154 & 0.7746 & 0.7671 & 0.7633 & 0.7532 & 0.7268 & 0.8253 & 0.8072 & 0.7482 & 0.7350 & 0.7445 \\
    EditTransfer~\cite{edittransfer} & 0.8275 & 0.8434 & 0.7853 & 0.7457 & 0.8081 & 0.8009 & 0.7314 & 0.8394 & 0.8605 & 0.8026 & 0.7594 & 0.8049 \\
    Prompt-Diffusion~\cite{prompt-diffusion} & 0.7723 & 0.7787 & 0.7038 & 0.6787 & 0.7134 & 0.6984 & 0.7001 & 0.7908 & 0.7645 & 0.7024 & 0.6913 & 0.7292 \\
    Cross~\cite{cross_image} & 0.8281 & 0.8332 & 0.7927 & 0.7553 & 0.8045 & 0.8004 & 0.7709 & 0.8554 & 0.8504 & 0.8048 & 0.7654 & 0.8068 \\
    SOER~\cite{aaai26} & \underline{0.8751} & \underline{0.8855} & \underline{0.8367} & \underline{0.8149} & \underline{0.8574} & \underline{0.8483} & \underline{0.7967} & \underline{0.8934} & \underline{0.8929} & \underline{0.8476} & \underline{0.8052} & \underline{0.8531} \\
    \rowcolor{blue!10}\methodtable & \textbf{0.8779} & \textbf{0.8875} & \textbf{0.8437} & \textbf{0.8276} & \textbf{0.8627} & \textbf{0.8562} & \textbf{0.8047} & \textbf{0.8937} & \textbf{0.8990} & \textbf{0.8573} & \textbf{0.8129} & \textbf{0.8590} \\
  \bottomrule
\end{tabular}
}

\noindent\textbf{SSIM (GT)}\\
\resizebox{\textwidth}{!}{%
\begin{tabular}{l|ccccccccccc|c}
  \toprule
  \textbf{Method} & \textbf{Face} & \textbf{Body} & \textbf{Biometric} & \textbf{Document} & \textbf{Contact} & \textbf{Identity} & \textbf{Financial} & \textbf{Monitor} & \textbf{License Plate} & \textbf{Time} & \textbf{Others} & \textbf{Overall} \\
  \hline
    VISII~\cite{visii} & 0.5470 & 0.7078 & 0.6592 & 0.6691 & 0.6846 & 0.6840 & 0.6531 & 0.7208 & 0.6672 & 0.6663 & 0.6649 & 0.6312 \\
    EditTransfer~\cite{edittransfer} & 0.7314 & 0.7101 & 0.6717 & 0.6257 & 0.7324 & 0.7270 & 0.6260 & 0.6898 & 0.7565 & 0.7156 & 0.6654 & 0.7116 \\
    Prompt-Diffusion~\cite{prompt-diffusion} & 0.5609 & 0.5641 & 0.4008 & 0.4238 & 0.4668 & 0.4401 & 0.3915 & 0.5015 & 0.5106 & 0.4248 & 0.4209 & 0.4813 \\
    Cross~\cite{cross_image} & 0.6723 & 0.6638 & 0.6552 & 0.6174 & 0.6789 & 0.6786 & 0.6184 & 0.6791 & 0.6748 & 0.6612 & 0.6299 & 0.6634 \\
    SOER~\cite{aaai26} & \underline{0.8291} & \underline{0.8091} & \underline{0.7568} & \underline{0.7239} & \underline{0.8203} & \underline{0.8138} & \underline{0.7041} & \underline{0.7976} & \underline{0.8368} & \underline{0.8038} & \underline{0.7436} & \underline{0.8016} \\
    \rowcolor{blue!10}\methodtable & \textbf{0.8308} & \textbf{0.8163} & \textbf{0.7684} & \textbf{0.7279} & \textbf{0.8345} & \textbf{0.8261} & \textbf{0.7172} & \textbf{0.8067} & \textbf{0.8443} & \textbf{0.8164} & \textbf{0.7571} & \textbf{0.8102} \\
  \bottomrule
\end{tabular}
}

\noindent\textbf{PSNR (GT)}\\
\resizebox{\textwidth}{!}{%
\begin{tabular}{l|ccccccccccc|c}
  \toprule
  \textbf{Method} & \textbf{Face} & \textbf{Body} & \textbf{Biometric} & \textbf{Document} & \textbf{Contact} & \textbf{Identity} & \textbf{Financial} & \textbf{Monitor} & \textbf{License Plate} & \textbf{Time} & \textbf{Others} & \textbf{Overall} \\
  \hline
    VISII~\cite{visii} & 15.4871 & 20.1307 & 20.8438 & 19.8133 & 20.4124 & 20.4850 & 19.5409 & 19.8902 & 19.3644 & 20.2288 & 19.6311 & 18.6009 \\
    EditTransfer~\cite{edittransfer} & 18.8927 & 18.5753 & 18.8239 & 17.3227 & 19.3264 & 19.2200 & 17.5987 & 17.8593 & 19.3296 & 19.2983 & 18.1614 & 18.8607 \\
    Prompt-Diffusion~\cite{prompt-diffusion} & 15.5054 & 16.1105 & 13.7255 & 13.9730 & 14.5920 & 14.2146 & 14.0056 & 14.9272 & 14.6380 & 14.4822 & 14.0136 & 14.7139 \\
    Cross~\cite{cross_image} & 18.2130 & 18.1453 & 18.9155 & 17.7239 & 19.0447 & 18.9969 & 18.1140 & 18.8917 & 18.1704 & 18.9260 & 17.9620 & 18.4962 \\
    SOER~\cite{aaai26} & \underline{23.1179} & \underline{22.8965} & \underline{21.9579} & \underline{20.6170} & \underline{23.3367} & \underline{23.1672} & \underline{20.2984} & \underline{21.6368} & \underline{23.5321} & \underline{23.1814} & \underline{21.2512} & \underline{22.7111} \\
    \rowcolor{blue!10}\methodtable & \textbf{23.4810} & \textbf{23.1770} & \textbf{22.7760} & \textbf{21.4012} & \textbf{24.3843} & \textbf{24.0594} & \textbf{21.2283} & \textbf{22.2339} & \textbf{24.0750} & \textbf{23.9854} & \textbf{21.9156} & \textbf{23.3774} \\
  \bottomrule
\end{tabular}
}
\end{table*}




Because applying an edit naturally shifts the output away from the original source, there is an inherent trade-off between source integrity and edit fidelity. Fig.~\ref{fig:tradeoff_recovery} visualizes this directly. VISII scores relatively high on source integrity but low on edit consistency, reflecting its tendency to under-edit. EditTransfer instead improves edit alignment (DirI 0.6817) but weakens source preservation (CLIP-Sim (Src) 0.6491). \method consistently occupies the upper-right region across all metric pairs, achieving the best balance between source preservation and edit transfer.

\begin{figure*}[t]
\centering
\includegraphics[width=\textwidth]{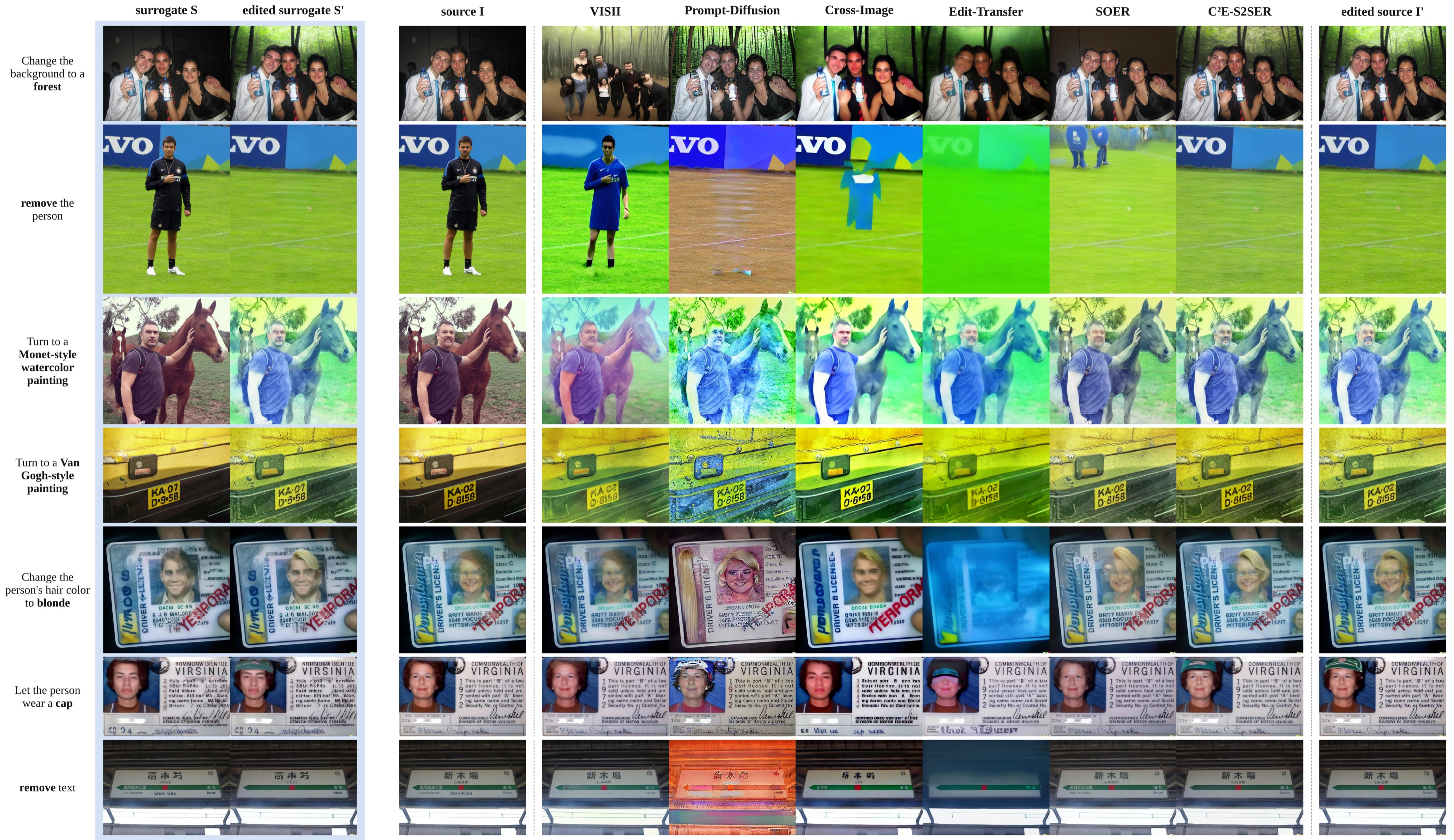}
\caption{Qualitative comparison for surrogate-to-source edit recovery. Each row shows the surrogate pair $(S,S')$, the private source image $I$, recovered results from different methods, and the  reference $I'$, covering diverse editing types and privacy categories.}
\label{fig:task2_qualitative}
\vspace{-10pt}
\end{figure*}

\subsubsection{Qualitative Results}

Fig.~\ref{fig:task2_qualitative} illustrates how different methods transfer the edit observed on the surrogate back to the private source image. The examples reveal several representative failure modes. For background replacement, the desired change should affect the scene behind the subjects while keeping the people and their relative layout unchanged. Some baselines change unrelated regions or introduce identity and lighting drift, whereas \method localizes the background change while preserving the source  structure. For object removal, the recovered result should remove the person and plausibly complete the field. Several baselines either leave residual human shapes, introduce unnatural color blocks, or erase surrounding context together with the target object, whereas \method produces a cleaner removal with more consistent background completion.

The style-transfer examples further show why recovery cannot rely on the text prompt alone. In the Monet-style and Van Gogh-style rows, the same instruction requires the model to infer the actual visual transformation from $(S,S')$, including the color palette and brush-like texture, and then apply it to the private source without replacing the source content. Several recovered results capture the overall edit tendency but lose fine-grained details around object boundaries, local textures, or source-specific structure. \method better preserves the original object geometry and applies a style change that is closer to the edited-source reference. The identity and text-related rows are more sensitive: changing hair color, adding a cap, or removing text should modify only the instructed attribute while keeping document layout and facial identity intact. Baselines frequently introduce spurious overlays, blur structured text regions, or change unrelated document details. \method shows fewer unintended modifications and maintains clearer source structure, supporting the quantitative finding that recovery should be evaluated jointly by source integrity, edit consistency, and edit-direction alignment.

\subsubsection{Generalization on InstructPix2Pix}
\begin{table*}[t]
\centering
\caption{Quantitative comparison on the InstructPix2Pix dataset. We adopt the same evaluation protocol to measure edit consistency and source integrity, assessing the generalization ability of each method on unseen images and prompts. VISII~\cite{visii} is excluded because it relies on a pretrained model trained on the InstructPix2Pix dataset. \textbf{Bold} denotes the best result and \underline{underline} denotes the second best result.}
\label{tab:summary_metrics}
\resizebox{\textwidth}{!}{%
\begin{tabular}{lcccccccc}
  \toprule
  \textbf{Method} & \textbf{CLIP-Sim (Src)} & \textbf{SSIM (Src)} & \textbf{PSNR (Src)} & \textbf{DirI} & \textbf{DirS} & \textbf{CLIP-Sim (GT)} & \textbf{SSIM (GT)} & \textbf{PSNR (GT)} \\
  \midrule
  Prompt-Diffusion~\cite{prompt-diffusion} & 0.7248 & 0.4300 & 11.7821 & 0.3699 & 0.1568 & 0.7417 & 0.4111 & 12.2141 \\
  EditTransfer~\cite{edittransfer} & 0.7684 & 0.5320 & 15.1055 & 0.3446 & 0.1717 & 0.7661 & 0.5251 & 15.9315 \\
  SOER~\cite{aaai26} & 0.7783 & 0.5413 & 15.1643 & 0.3817 & 0.2077 & 0.7857 & 0.5340 & 16.0415 \\
  Cross~\cite{cross_image} & \underline{0.8334} & \underline{0.6135} & \underline{17.0173} & \underline{0.3970} & \underline{0.2193} & \underline{0.8231} & \underline{0.5850} & \underline{16.9894} \\
  \rowcolor{blue!10}\methodtable & \textbf{0.8520} & \textbf{0.6626} & \textbf{18.1497} & \textbf{0.4774} & \textbf{0.2342} & \textbf{0.8595} & \textbf{0.6449} & \textbf{18.1600} \\
  \bottomrule
\end{tabular}
}
\end{table*}
Table~\ref{tab:summary_metrics} evaluates generalization on InstructPix2Pix, where images and prompts are outside SPPE. \method achieves the best result on all eight metrics. On source integrity, Cross-Image Attention provides the second-best performance, which is expected given its zero-shot design that requires no training on any specific distribution. The fact that \method, fine-tuned on SPPE, matches or exceeds this on unseen data suggests that surrogate-pair conditioning provides a transferable signal rather than overfitting to SPPE-specific patterns. On edit consistency and directional metrics, \method shows a clearer advantage over all baselines, indicating stronger edit transfer under distribution shift.

\subsubsection{Ablation Study}
\begin{table*}[t]
\centering
\caption{Ablation study. \textbf{Bold} = best, \underline{underline} = second best within each group.}
\label{tab:ablation}
\resizebox{\textwidth}{!}{%
\begin{tabular}{llcccccccc}
  \toprule
  \textbf{Group} & \textbf{Method} & \textbf{CLIP-Sim (Src)} & \textbf{SSIM (Src)} & \textbf{PSNR (Src)} & \textbf{DirI} & \textbf{DirS} & \textbf{CLIP-Sim (GT)} & \textbf{SSIM (GT)} & \textbf{PSNR (GT)} \\
  \midrule
  \multirow{4}{*}{\textbf{Iteration}} & Iter 5000 & 0.6956 & 0.6594 & 16.9870 & 0.6848 & 0.5969 & 0.8227 & 0.7510 & 21.1782 \\
   & Iter 6000 & 0.6946 & 0.6621 & 17.2635 & \textbf{0.7411} & \textbf{0.6719} & \textbf{0.8590} & \textbf{0.8102} & \textbf{23.3774} \\
   & Iter 7000 & \underline{0.6962} & \underline{0.6645} & \underline{17.4038} & \underline{0.7166} & \underline{0.6377} & \underline{0.8441} & \underline{0.7949} & \underline{23.3581} \\
   & Iter 8000 & \textbf{0.7193} & \textbf{0.6778} & \textbf{17.6723} & 0.6813 & 0.5910 & 0.8270 & 0.7571 & 21.0141 \\
  \midrule
  \multirow{3}{*}{\textbf{Module}} & Base & 0.6590 & 0.6244 & 15.9569 & 0.6992 & 0.6212 &  0.8190 &0.7469&  20.3887\\
   & + Cycle & \underline{0.6854} & \underline{0.6459} & \underline{16.6804} & \underline{0.7183} & \underline{0.6431} & \underline{0.8412} & \underline{0.7818} & \underline{21.2121} \\
   & + Edit Tag & \textbf{0.6946} & \textbf{0.6621} & \textbf{17.2635} & \textbf{0.7411} & \textbf{0.6719} & \textbf{0.8590} & \textbf{0.8102} & \textbf{23.3774} \\
  \bottomrule
\end{tabular}
}
\end{table*}

Table~\ref{tab:ablation} studies two factors: training schedule and model components. For the training schedule, we adopt a staged strategy: the first 3000 iterations optimize forward recovery with edit-conditioned tag guidance, while subsequent training further introduces cycle-consistent recovery regularization. Source-reference metrics measure preservation quality, whereas ground-truth-reference and directional metrics evaluate edit fidelity. Introducing cycle consistency with a moderate training length improves both preservation and edit fidelity, suggesting that the regularizer helps suppress over-editing by encouraging the recovered result to remain compatible with the source domain.  The model trained with the full staged schedule achieves the best performance on all edit-consistency and GT-reference metrics. When training is further prolonged beyond this setting, source-reference SSIM and PSNR continue to increase slightly, with PSNR (Src) rising from 17.26 to 17.67  at 8000 iterations. However, edit-direction consistency decreases, with DirI dropping from 0.7411 to 0.6813. This suggests that excessive training with cycle regularization makes the model increasingly biased toward source appearance, improving preservation but weakening edit transfer. Therefore, the full staged schedule provides the best balance between source integrity and edit fidelity.

For model components, each addition brings consistent gains across both preservation and consistency metrics. The cycle-consistent recovery regularizeris introduced after the forward warm-up and improves over the base model on all metrics, confirming that reverse reconstruction constrains unnecessary modifications without suppressing edit transfer. Further incorporating the weakly supervised edit-conditioned tag guidance  yields the strongest overall performance across all metrics. The directional metrics DirI and DirS improve from 0.7183 to 0.7411 and from 0.6431 to 0.6719, respectively, while PSNR (GT) shows a substantial gain from 21.21 to 23.38 dB.
 These improvements suggest that  edit categorization provides a useful structural signal that discourages over-reliance on the text prompt and encourages the model to ground its edit interpretation in the  transformation observed in the surrogate pair. 

\subsection{Directional Label Validation}

\begin{figure}[t]
\centering
\includegraphics[width=\linewidth]{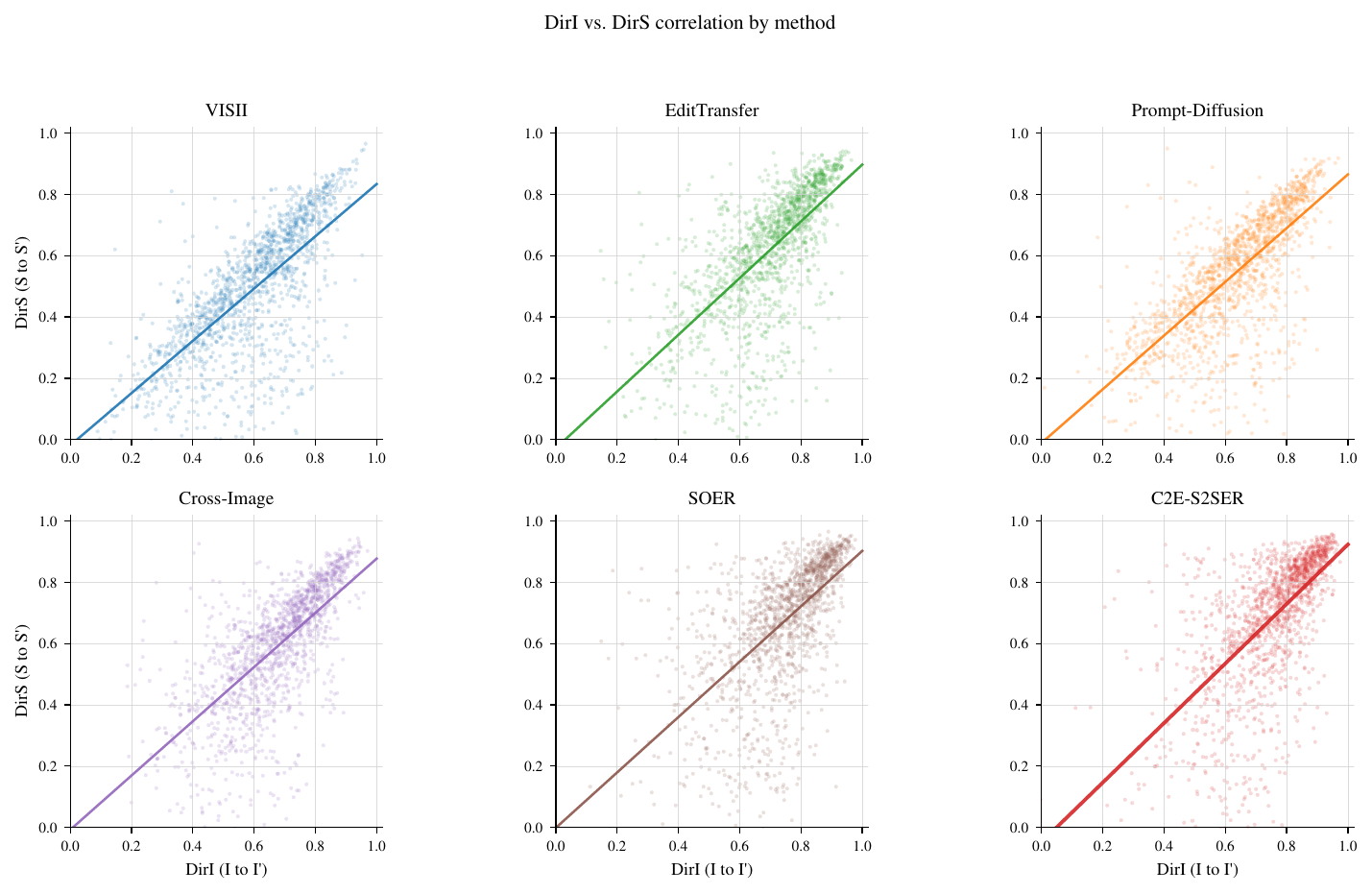}
\caption{Correlation between source-domain and surrogate-domain edit-direction consistency across recovery methods.}
\label{fig:diri_dirs_correlation}
\end{figure}
Fig.~\ref{fig:diri_dirs_correlation} provides an empirical sanity check for the editability label used in Task 1. Across representative recovery methods, DirI and DirS show a consistent positive correlation. In other words, samples whose recovered edits better follow the surrogate-domain edit direction $(S \rightarrow S')$ also tend to better follow the source-domain edit direction $(I \rightarrow I')$. This trend indicates that the surrogate editing direction is not merely an auxiliary visual cue, it carries edit information that is transferable to the private source domain.

This observation supports the use of source-surrogate editing-direction consistency as a pseudo label for surrogate editability. If a surrogate preserves the semantics needed for the instruction, the MLLM edit on the surrogate should move in a direction similar to the edit on the original image. Conversely, when the surrogate loses prompt-relevant content, the surrogate-domain edit direction becomes less aligned with the source-domain edit direction, making the edited surrogate less reliable for recovery. The positive DirI--DirS relation therefore connects the Task 1 label definition with the downstream recovery objective.


\section{Conclusion}

This paper studies surrogate-driven MLLM image editing for privacy-sensitive scenarios, where a locally generated surrogate is edited by a cloud-side MLLM and the observed edit is then recovered on the original source image. We identify surrogate-to-source recovery as a largely unexplored yet essential problem in this setting. To support systematic evaluation, we introduce SPPE, a benchmark dataset with fine-grained privacy categories, diverse editing instructions, region-level annotations, surrogate editing pairs, and edited-source references. SPPE defines two complementary tasks: editability assessment, which estimates before editing whether a surrogate can support consistent edit behavior, and surrogate-to-source edit recovery, which evaluates whether the edited surrogate can be transferred back to the private source image.
Based on these tasks, we propose ERMA for instruction-aware editability assessment and \method for surrogate-to-source edit recovery, which uses the surrogate editing pair as visual edit evidence together with edit-conditioned guidance and cycle-consistent regularization.
Experiments on SPPE show that ERMA predicts surrogate editability more accurately than representative image quality assessment baselines. For surrogate-to-source recovery, \method achieves stronger source integrity and  edit consistency on SPPE, and further shows robust generalization on InstructPix2Pix. 


\bibliography{sn-bibliography}

\end{document}